\documentclass{article} %
\usepackage{iclr2024_conference,times}

\usepackage{hyperref}
\usepackage{url}
\usepackage{comment}
\usepackage[utf8]{inputenc}
\usepackage{amsmath}
\usepackage{amssymb}
\usepackage{amsmath}
\usepackage{amsthm}
\usepackage{pgfplots}
\usepackage{graphicx}
\usepackage{booktabs}
\usepackage{bbm}
\usepackage{accents}
\usepackage{float}
\usepackage{enumitem}
\usepackage{caption}
\usepackage{subcaption}
\usepackage{adjustbox}
\usepackage{thm-restate}
\usepackage{xcolor}
\usepackage{wrapfig}

\definecolor{mpl_red}{HTML}{D62728}
\definecolor{mpl_blue}{HTML}{1F77B4}

\hypersetup{
    colorlinks,
    linkcolor={red!50!black},
    citecolor={blue!50!black},
    urlcolor={blue!80!black}
}

\newtheorem{theorem}{Theorem}
\numberwithin{theorem}{section}
\newtheorem{definition}[theorem]{Definition}

\newtheorem{corollary}[theorem]{Corollary}

\newtheorem{example}[theorem]{Example}
\newtheorem{proposition}[theorem]{Proposition}

\newcommand{\EE}{\mathbb{E}}
\newcommand{\Prob}{\mathbb{P}}
\newcommand{\Var}{\mathrm{Var}}

\newcommand{\oracle}[1]{{O_{#1}}}
\newcommand{\utility}{u}
\newcommand{\exutility}{\bar{\utility}}
\newcommand{\learnedutility}{\hat{\utility}}
\newcommand{\loss}{L}
\newcommand{\noise}{\epsilon}
\newcommand{\bordacount}{\text{BC}}
\newcommand{\altspace}{\mathcal{A}}
\newcommand{\alta}{a}
\newcommand{\altb}{b}
\newcommand{\altc}{c}
\newcommand{\unseenspace}{\mathcal{Z}}
\newcommand{\unseen}{z}
\newcommand{\unseendist}{\mathcal{D}_\unseen}
\newcommand{\comparisonprob}{p}
\newcommand{\btlprob}{\comparisonprob^\text{BTL}}
\newcommand{\uniformdist}{\text{Unif}}
\newcommand{\bernoulli}{\mathcal{B}}
\newcommand{\learneddist}{\smash{\hat{\mathcal{D}}}}

\newcommand{\smallparagraph}[1]{\noindent\textbf{#1}\quad}

\makeatletter
\newcommand{\ssymbol}[1]{^{\@fnsymbol{#1}}}
\makeatother

\title{Distributional Preference Learning: \\ Understanding and Accounting for Hidden Context in RLHF}

\author{Anand Siththaranjan \thanks{Equal contribution.} \qquad Cassidy Laidlaw $\ssymbol{1}$ \\
University of California, Berkeley \\
\texttt{\{anandsranjan,cassidy\_laidlaw\}@cs.berkeley.edu} \\
\And
Dylan Hadfield-Menell \\
Massachusetts Institute of Technology \\
\texttt{dhm@csail.mit.edu} \\
}

\iclrfinalcopy %
\begin{document}

\maketitle

\begin{abstract}
In practice, preference learning from human feedback depends on incomplete data with hidden context. Hidden context refers to data that affects the feedback received, but which is not represented in the data used to train a preference model. This captures common issues of data collection, such as having human annotators with varied preferences, cognitive processes that result in seemingly irrational behavior, and combining data labeled according to different criteria. We prove that standard applications of preference learning, including reinforcement learning from human feedback (RLHF), implicitly aggregate over hidden contexts according to a well-known voting rule called \emph{Borda count}. We show this can produce counter-intuitive results that are very different from other methods which implicitly aggregate via expected utility. Furthermore, our analysis formalizes the way that preference learning from users with diverse values tacitly implements a social choice function. A key implication of this result is that annotators have an incentive to misreport their preferences in order to influence the learned model, leading to vulnerabilities in the deployment of RLHF. As a step towards mitigating these problems, we introduce a class of methods called \emph{distributional preference learning} (DPL). DPL methods estimate a distribution of possible score values for each alternative in order to better account for hidden context. Experimental results indicate that applying DPL to RLHF for LLM chatbots identifies hidden context in the data and significantly reduces subsequent jailbreak vulnerability.
Our code and data are available at \url{https://github.com/cassidylaidlaw/hidden-context}.
\end{abstract}

\section{Introduction}
Encoding human preferences and values into interactive learning systems is an essential component for making those systems safe and socially beneficial. To accomplish this, modern machine learning models, such as large language model (LLM) chatbots like ChatGPT and Claude, are trained with feedback from human evaluators. This method, often called reinforcement learning from human feedback (RLHF), seeks to align system behavior with the preferences of annotators. In this paper, we study how RLHF infers preferences when there is \emph{hidden context} that influences human evaluations.

Hidden context is any information that affects preference annotations but is not given as input to the learned utility or reward model. It can arise through several mechanisms. For instance, when feedback is collected from many different people, annotator identity is hidden context: it affects the annotations, since different annotators could have very different preferences, but it is not input to the reward model, since the annotators' data is combined anonymously. Other sources of hidden context include human irrationality and evaluation according to multiple objectives.

To motivate the consequences of naive preference learning with hidden context, consider the following hypothetical scenario:
\begin{example}
    \label{ex:college_admissions}
    A company has developed an AI assistant to help high school students navigate college admissions. They implement RLHF by asking their customers for feedback on how helpful the chatbot's responses are.
    Among other questions, this process asks users whether or not they prefer to see information about the Pell Grant, an aid program for low-income students.
    Because the population of customers is biased towards high-income students, most feedback indicates that users prefer other content to content about the Pell Grant. As a result, RLHF trains the chatbot to provide less of this kind of information. This marginally improves outcomes for the majority of users, but drastically impacts lower-income students, who rely on these recommendations to understand how they can afford college.
\end{example}

\begin{figure}
    \centering
    \includegraphics[clip, trim=0in 0.8in 0in 0in, page=1]{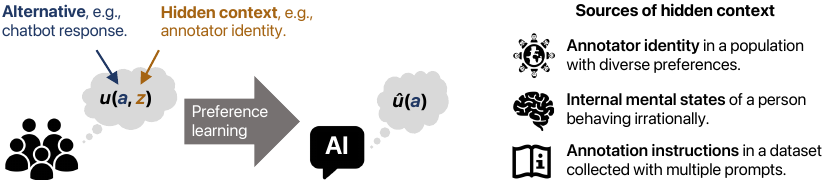}
    \caption{We analyze the effects of \emph{hidden context} on preference learning, which is one of the key steps in reinforcement learning from human feedback (RLHF). Hidden context is any information that affects the annotator's assessment of the utility of different alternatives, but is not input to the learned utility or reward model. Our framework emcompasses many potential issues with preference learning, including human irrationality, diverse preferences among annotators, and combining multiple objectives (Section \ref{sec:setup}). We prove that preference learning implicitly aggregates over hidden context using a rule called \emph{Borda count} (Section \ref{sec:perspectives}).}
    \label{fig:aggregation}
\end{figure}

The heart of this issue is that common preference learning approaches assume that all relevant features are provided as input to the reward model. However, when there is hidden context---which is almost always the case---this assumption is false. As a result, standard methods can have unexpected and undesirable consequences. In Example \ref{ex:college_admissions}, relevant context about the annotator's identity (i.e. their income level) is missing from the data. The implicit aggregation over preferences biases the outcome in favor of high-income applicants. In this work, we take steps to better understand the implications of unobserved context in preference learning and consider technical approaches to identify when such situations occur.

In Section~\ref{sec:setup} we present a formal model of preference learning with hidden context. We show that our model can represent many challenges in preference learning, such as combining data from different users, accounting for irrationality, and optimizing for multiple objectives. Since these challenges are ubiquitous, understanding their implications is crucial for safely deploying RLHF-trained models. 

In Section~\ref{sec:perspectives}, we use our model to develop theoretical results on the consequences of hidden context in preference learning.
First, we provide a precise characterization of the utility function that preference learning will output when there is hidden context. In particular, we show that preference learning implicitly aggregates over hidden context using a rule called the \emph{Borda count}. We explore the implications of this finding, identifying cases when Borda account aggregates preferences in unintuitive ways quite different from other methods like regression. Furthermore, when data is combined from many annotators, preference learning implicitly defines a \emph{social welfare functional} that aggregates their preferences. We use existing results from the social choice literature to expose another problem arising from hidden context: annotators may have an incentive to misreport their preferences to influence the learned reward function.

Next, we consider the design of preference learning methods that more gracefully account for hidden context. %
In Section \ref{sec:mitigate}, we propose \emph{distributional preference learning} (DPL). DPL estimates a distribution over utility values for each input instead of a single real-valued output. This allows the method to detect situations where unobserved context could influence preferences. We show how DPL can detect the effects of missing features through an explained variance ($r^2$) metric.

We validate DPL in two ways. First, we conduct a small-scale synthetic experiment with a 1-dimensional space of alternatives that allows us to directly compare to Borda count. Next, we apply DPL to a real-world dataset of preferences for use in RLHF. In this case, the preference data is collected according to two distinct objectives. In one subset of the data, raters were asked to prefer helpful and honest responses. In the other subset, raters were asked to prefer responses that did not respond to harmful requests. %
This introduces hidden context because the single reward model is trained on the combined data.
We find that DPL is able to identify this hidden context automatically and identifies the uncertainty when these competing goals are at odds. 

Beyond identifying potential instances of relevant hidden context, our experiments indicate that DPL can be used to develop guardrails that protect against jailbreaks. \citet{wei_jailbroken_2023} showed that many jailbreaks succeed by pitting the helpfulness and harmlessness objectives of chatbots against one another. This means that some jailbreaks can be understood as a consequence of hidden context. As a result, it is possible to detect this class of jailbreaks by leveraging the distribution of utilities we get from DPL. In particular, risk-aversion with respect to the distribution of learned utilities can dramatically reduce the rate at which the preference model prefers jailbroken responses. This is because DPL models capture the disagreement in the training dataset between the two objectives. Thus, risk-aversion penalizes responses that cause the two objectives to diverge.

We summarize our contributions as follows:
\begin{enumerate}
    \item we identify and formally characterize the problem of preference learning with hidden context, and describe a number of settings where it may arise;
    \item we show that preference learning with hidden context implicitly implements Borda count, which can have counter-intuitive implications and introduce incentives for annotators to misrepresent their preferences;
    \item we introduce distributional preference learning and show that it can detect and mitigate some effects of hidden context in LLM-based preference models.
\end{enumerate}

\section{Setting and Related Work}
\label{sec:setup}

We begin by formally describing the problem of preference learning with hidden context. We first review the standard preference learning framework and then introduce our extension that accounts for hidden context.

In standard preference learning, the goal is to estimate an unknown utility function $\utility: \altspace \to \mathbb{R}$ which assigns a utility value to each of a finite set of alternatives $\altspace$. For instance, in the case of a chatbot, the alternatives could be the possible responses to a prompt, and the utility function would describe how much a particular response is preferred.
To estimate $\utility$, we observe the outcome of comparisons between pairs of alternatives $(\alta, \altb)$. We assume there is a fixed probability for any pair of alternatives $(\alta, \altb)$ that $\alta$ will be preferred to $\altb$; we denote this probability $\comparisonprob_\utility(\alta, \altb)$ and assume that $\comparisonprob_\utility(\alta, \altb) + \comparisonprob_\utility(\altb, \alta) = 1$; that is, the order in which the alternatives are presented does not matter.
In the ideal case, comparison outcomes would exactly reflect the utility function, i.e., $\comparisonprob_\utility(\alta, \altb) = \textbf{1}\{\utility(\alta) > \utility(\altb)\}$.
Realistically, however, preference comparison data never exactly follows a single utility function. To account for the fact that people are noisy and/or inconsistent in their feedback, a common assumption is that instead preference comparisons are made according to a Bradley-Terry-Luce (BTL) model \citep{rajkumar_statistical_2014}, also sometimes known as Boltzmann-rational model \citep{jeon_reward-rational_2020}:
\begin{equation}
\label{eq:bradley_terry}
    \btlprob_\utility(\alta, \altb) = \frac{e^{\utility(\alta)}}{e^{\utility(\alta)} + e^{\utility(\altb)}}.
\end{equation}
In (\ref{eq:bradley_terry}), the higher $\utility(\alta)$ is compared to $\utility(\altb)$, the more likely the outcome of the comparison is to prefer $\alta$ to $\altb$; as the utilities for $\alta$ and $\altb$ are closer, the comparison outcome moves towards uniformly random.

The most commonly used method for estimating the utility function $\utility$ from preference data is to fit the maximum likelihood estimator (MLE) under the BTL model in (\ref{eq:bradley_terry}).
To derive the MLE, we consider the limit of infinite data and assume that preference comparisons are elicited for uniformly randomly selected pairs of alternatives. The MLE for the utility function $\learnedutility$ is given by minimizing the following loss function:
\begin{align}
    \label{eq:unregularized_optimization}
    \learnedutility & = \arg\min_{\learnedutility} \loss(\learnedutility; \utility) \\
    & = \arg\min_{\learnedutility} \frac{1}{|\altspace| (|\altspace| - 1)} \sum_{\alta \neq \altb} - \comparisonprob_\utility(\alta, \altb) \log \left( \frac{e^{\learnedutility(\alta)}}{e^{\learnedutility(\alta)} + e^{\learnedutility(\altb)}} \right) - \left( 1 - \comparisonprob_\utility(\alta, \altb) \right) \log \left( \frac{e^{\learnedutility(\altb)}}{e^{\learnedutility(\alta)} + e^{\learnedutility(\altb)}} \right). \nonumber
\end{align}
Although $\learnedutility$ might be chosen from a parametric class like a neural network, we assume for theoretical purposes that $\loss(\learnedutility; \utility)$ is optimized over \emph{all} possible $\learnedutility: \altspace \to \mathbb{R}$. 
In some cases, $\loss$ may not have any minimum, so we consider a regularized version of (\ref{eq:unregularized_optimization}) where $\lambda > 0$ is a tunable coefficient:
\begin{equation}
    \label{eq:regularized_optimization}
    \learnedutility = \arg\min_{\learnedutility} \loss(\learnedutility; \utility) + \frac{\lambda}{2} \sum_{\alta \in \altspace} \learnedutility(\alta)^2.
\end{equation}
The optimization objective in (\ref{eq:regularized_optimization}) is strongly convex and thus has a unique global minimum; see Appendix \ref{sec:loss_is_convex} for a proof.

\subsection{Hidden Context}
While preference learning based on (\ref{eq:regularized_optimization}) has been widely deployed and enjoyed some success, it rests on assumptions that often do not hold in practice. In particular, irrationality, partial observability, and diversity of preferences among a population all challenge the BTL model on which the usual preference learning loss is based. We argue that all of these cases can be understood as special cases of a general phenomenon: \textbf{hidden context}. For concreteness, consider again Example \ref{ex:college_admissions}. The key problem in the example is a mismatch between the information that influences the user's feedback and the information that the preference learning algorithm uses to estimate utilities based on that feedback. The user gives feedback that depends on their financial situation, while the learned utility model observes request-response pairs. Thus, the preference learning algorithm must produce a single ordering over alternatives that implicitly aggregating feedback over the hidden context of whether the user is high- or low-income.

To model hidden context in preference learning, we extend the preference learning formalization to utility functions $\utility: \altspace \times \unseenspace \to \mathbb{R}$ over a space of observed features $\alta \in \altspace$ and hidden context $\unseen \in \unseenspace$. Let $\unseendist$ be a distribution over $\unseenspace$.
In Example \ref{ex:college_admissions}, $\unseen \in \{0, 1\}$ could represent whether the user is low- or high-income; then perhaps $\unseen \sim \bernoulli(0.8)$ if 80\% of users are high-income (where $\bernoulli(p)$ represents a Bernoulli random variable with mean $p$).
Given $\utility(\alta, \unseen)$ and $\unseendist$, we can calculate the probability that one alternative $\alta$ is chosen over another $\altb$ given that $\unseen$ is hidden:
\begin{equation}
    \label{eq:unseen_comparison_prob}
    \comparisonprob_{\utility, \unseendist} (\alta, \altb) = \EE_{\unseen \sim \unseendist} \left[ \oracle{\utility}(\alta, \altb, \unseen) \right] \quad \text{where} \quad \oracle{\utility}(\alta, \altb, \unseen) = \begin{cases}
        1/2 \quad \text{if } \utility(\alta, \unseen) = \utility(\altb, \unseen) \\
        \mathbf{1} \{ \utility(\alta, \unseen) > \utility(\altb, \unseen) \} \quad \text{o.w.}
    \end{cases}
\end{equation}
$\comparisonprob_{\utility, \unseendist}$ marginalizes over the distribution of the hidden context $\unseen$ and thus reflects the comparison data available to the preference learning algorithm. Our model of hidden contexts can represent many settings where preference learning is difficult:

\smallparagraph{Partial observability.} There may be variables that are observable by the human making preference comparisons but not by the AI system, which learns from that data. For instance, suppose annotators' preferences depend on the day of the week or the month of the year, but the estimated utility function ignores the date the comparisons were made.

\smallparagraph{Multiple objectives.} System designers may combine data about user preferences over multiple, different objectives. For instance, the Anthropic HH-RLHF dataset \citep{bai_training_2022} contains one subset with comparisons of chatbot responses based on harmlessness and another subset with comparisons based on helpfulness. When these subsets are combined, the objective that was used to make the comparison (in this case, either harmlessness or helpfulness) is a hidden context. We explore this case more in Section~\ref{sec:experiments}.

\smallparagraph{Population with diverse preferences.} Preference learning is almost always applied to data aggregated from many annotators who may have very different utility functions (e.g., \citet{bai_training_2022} observe high intra-annotator disagreement). If $\unseen$ represents the annotator who makes a comparison, then $\utility(\cdot, \unseen)$ could represent the utility function for that annotator. However, when the data is used to train a single utility function $\learnedutility(\cdot)$, then the annotator's identity $\unseen$ is a hidden context.

\smallparagraph{Irrational and noisy decisions.} Various types of irrationality could be modeled as unseen latent variables that affect a person's decision-making. For instance, to represent a person making noisy utility estimates, one could let $\unseenspace = \mathbb{R}^{|\altspace|}$, $\unseen(\alta) \overset{\text{iid}}{\sim} \mathcal{N}(0, 1)$, and $\utility(\alta, \unseen) = \mu(\alta) + \unseen(\alta)$ for some $\mu : \altspace \to \mathbb{R}$. That is, the person has an underlying utility $\mu(\alta)$ for each alternative but makes comparisons based on that utility plus independently sampled Gaussian noise representing irrationality in their utility assessments. This is equivalent to the Thurstone-Mosteller model of noisy decision making \citep{handley_comparative_2001}.

Due to the ubiquity of these settings, preference learning is nearly always performed with hidden context. This means that the learned utility function $\learnedutility(\alta)$, which only depends on the seen features $\alta$, must somehow aggregate over the hidden contexts $\unseen$. We aim to understand and mitigate the consequences of this ubiquitous challenge.

\subsection{Related Work}

Preference learning and its use in reinforcement learning have a long history \cite{akrour_april_2012,busa-fekete_survey_2014,sadigh_active_2017,christiano_deep_2017,pacchiano_dueling_2021}. 
As part of RLHF, preference learning has been widely used recently for training large language models (LLM) to give outputs according to human preferences \citep{ziegler_fine-tuning_2020,stiennon_learning_2020,askell_general_2021,bai_training_2022,bai_constitutional_2022,ouyang_training_2022}. It has also been extensively analyzed in theory; some results focus on its sample complexity in various settings \citep{chen_spectral_2015,shah_estimation_2015,shah_simple_2018,heckel_approximate_2018,hendrickx_minimax_2020,chambers_recovering_2021} or other directions such as the statistical identifiability of preferences \citep{zhao_learning_2020,skalse_invariance_2023}, the computational efficiency of preference learning \citep{maystre_fast_2015}, Bayesian preference learning \citep{caron_efficient_2010}, or the combination of preference learning and reinforcement learning \citep{zhu_principled_2023}. However, to our knowledge, no prior work has specifically analyzed the behavior of preference learning with hidden context.

The challenges of preference learning that we group as cases of ``hidden context'' have also been studied individually. There has been some work on explicitly modeling annotator disagreement \citep{fleisig_when_2023,baumler_which_2023} as well as other approaches to learning from annotators with diverse preferences \citep{jia_embedding_2023,dumoulin_density_2023,mishra_ai_2023,fish_generative_2023}. Other work has studied the effects of human irrationality or non-BTL models of human behavior on preference learning \citep{bobu_less_2020,lee_b-pref_2021,laidlaw_uncertain_2021,knox_models_2022,laidlaw_boltzmann_2022}, which under our framework can be modeled as hidden context. \citet{zhuang_consequences_2020} and \citet{dai_safe_2023} study the optimization of multiple objectives learned from human preferences. Finally, related to our connections with social choice theory in Section \ref{sec:perspectives}, some previous work has associated preference or reward learning with concepts in economics, such as voting rules \citep{conitzer_common_2005}, incentive compatibility \citep{echenique_incentive_2019}, and mechanism design \citep{fickinger_multi-principal_2020}.

\section{Theoretical Analysis}
\label{sec:perspectives}

In this section, we provide a theoretical analysis of standard preference learning---formally known as the BTL estimator---in the presence of hidden context. We are particularly interested in the question of how preference learning estimates a utility function $\learnedutility(\alta)$ which only depends on the observed alternatives $\alta$, despite using data that is generated from a utility function $\utility(\alta, \unseen)$ which depends both on $\alta$ and hidden context $\unseen$. In order to do this, preference learning must implicitly aggregate over hidden context to estimate a single utility value for each alternative $\alta$ (Figure \ref{fig:aggregation}).

We study this implicit aggregation from two perspectives: identification of expected utilities and social choice theory. In the former case, we are interested in when preference learning will aggregate over hidden context via expected value, which is a natural and often desirable aggregation method. In the latter case, we consider the setting where the hidden context is an annotator's identity. In this case, we show that preference learning acts as a \emph{social welfare functional}: a method for constructing a societal utility function from many individual utilities.

We begin our analysis by precisely describing the behavior of preference learning with hidden context. In particular, we can show that a utility function $\learnedutility(\alta)$ learned with the BTL loss as in (\ref{eq:regularized_optimization}) implicitly aggregates utilities over the hidden contexts $\unseen$ using a rule called \emph{Borda count}. We define the Borda count $\bordacount(\alta)$ of an alternative $\alta$ as $\bordacount(\alta) = \smash{\frac{1}{| \altspace |}} \sum_{\altb \in \altspace} \comparisonprob_{\utility, \unseendist}(\alta, \altb)$.
That is, the Borda count is the average probability that the alternative is preferred to other alternatives. If an alternative is almost always preferred to every other alternative, then its Borda count will be close to 1; if it is almost always dispreferred, the Borda count will be near 0. We use the term Borda count as a reference to the well-known voting rule of the same name---a connection we expand on in Section \ref{sec:sct}.
\begin{restatable}{theorem}{thmlearntnoise}
    \label{theorem:learnt-noise}
    BTL preference learning implicitly aggregates hidden context according to Borda count.
    That is, if $\learnedutility$ is optimized according to (\ref{eq:regularized_optimization}), then
    $\forall \alta, \altb \in \altspace$, $\learnedutility(\alta) > \learnedutility(\altb)
        \Leftrightarrow
        \bordacount(\alta) > \bordacount(\altb)$.
\end{restatable}
We defer all proofs to Appendix \ref{sec:proofs}. According to Theorem \ref{theorem:learnt-noise}, the learned utility function and Borda count differ by only a monotonic transformation. If we use reinforcement learning or another optimization technique to search for the alternative $\alta$ which maximizes $\learnedutility(\alta)$---as one does in RLHF---then the optimal alternative will the same as that which maximizes the Borda count $\bordacount(\alta)$. Similar results that relate preference learning and Borda count were previously explored by \citet{rajkumar_statistical_2014}, although they do not consider the setting of hidden context.

While Theorem \ref{theorem:learnt-noise} precisely describes the results of preference learning with hidden context, its implications are unclear. Is Borda count a useful way of aggregating over hidden contexts in practice, and how does it compare to other aggregation rules? To answer this question, we give multiple perspectives on preference learning with hidden context using the result of Theorem \ref{theorem:learnt-noise}. First, we compare preference learning to least-squares regression with hidden context.
Then, we analyze learning from a population with diverse preferences through the lens of social choice theory.

\subsection{Comparison to expected utility and least-squares regression}
One desirable property of preference learning with hidden context would be if it converged to the \emph{expected utility} for each alternative when marginalizing over hidden context, which we denote by $\exutility(\alta) = \EE_{\unseen \sim \unseendist}[\utility(\alta, \unseen)]$. For instance, one can show that least-squares utility \emph{regression} converges to the expected utility when there is hidden context; see Appendix \ref{sec:least_squares_robust} for a formal statement and proof.

The fact that least-squares utility regression converges to $\learnedutility = \exutility$ shows that, in some sense, it gracefully degrades in the presence of hidden context.
Although there are problems with maximizing expected utility, it is a well-understood method of aggregating utilities over hidden contexts that has desirable decision-theoretic properties. Thus, it would be helpful if the utility function $\learnedutility(\alta)$ learned by preference learning with hidden context were equivalent to the expected utility $\exutility(\alta)$. In this section, we characterize when the output of \emph{preference learning} with hidden context is equivalent to that of \emph{utility regression}. We identify sufficient conditions for the two to be equivalent and present cases where they differ. 

\smallparagraph{Positive results}
In some cases, we can show that preference learning does identify a utility function that is equivalent to the expected utility. Our result requires that the zero-mean ``noise'' induced by hidden context is identical across alternatives and reasonably distributed. Specifically, denote by $\noise(\alta) = \utility(\alta, \unseen) - \exutility(\alta)$ (where $\unseen \sim \unseendist$) to be the random variable representing the residual utility of an alternative $\alta$ after subtracting its expected utility.
\begin{restatable}{theorem}{thmidentifiabilitysupportaroundzero}
\label{theorem:identifiability-support-around-zero}
    Let $\noise(\alta)$ be independent and identically distributed for all $\alta \in \altspace$. Furthermore, suppose $\epsilon(\alta) - \epsilon(\altb)$ has support around 0, i.e., $\forall \delta > 0$, $F_{\alta, \altb}(\delta) > F_{\alta, \altb}(0) = \frac{1}{2}$, where $F_{\alta, \altb}$ is the cumulative distribution function of $\noise(\alta) - \noise(\altb)$. Then the utility function $\learnedutility$ learned by minimizing (\ref{eq:regularized_optimization}) satisfies $\learnedutility(\alta) > \learnedutility(\altb)
        \Leftrightarrow
        \exutility(\alta) > \exutility(\altb)$
    for any $\alta, \altb \in \altspace$.
\end{restatable}
Many noise distributions, such as uniform and normal distributions, clearly satisfy the assumptions of Theorem \ref{theorem:identifiability-support-around-zero}. Thus, as long as the noise caused by hidden context does not vary across alternatives and is not too unusual, we generally expect that preference learning will give a utility function with the same ordering over alternatives as the expected utility. This means that it performs similarly to least-squares regression.

\smallparagraph{Negative results}
In other cases, preference learning can behave quite differently from utility regression. Example \ref{ex:college_admissions} describes such a case. The expected utility of telling students about Pell Grants is higher than the expected utility of not telling them, since it is of great benefit to low-income students and only small inconvience to high-income students. However, the Borda count is lower since the high-income majority prefer not to hear about the grants. This results in preference learning assigning higher utility to \emph{not} giving the grant information, while regression would assign higher utility to giving it.

One might suppose that preference learning and regression disagree in this case because the majority of users prefer the alternative with lower expected utility, and preference learning gives a learned utility function which assigns higher utilities to alternatives preferred to by the majority of users. As long as the majority of feedback agrees with the ordering given by the expected utility, will preference learning and regression give the same result? The following theorem shows that this is not the case.
\begin{restatable}{proposition}{proppairwisemajority}
    \label{proposition:pairwise_majority}
    $\exists\altspace, \unseendist,\utility$ s.t $\forall \alta, \altb \in \altspace$, $\left [ \exutility(\alta) > \exutility(\altb)\right ] \Rightarrow  \left [ \comparisonprob_{\utility, \unseendist}(\alta, \altb) > 1/2\right ]$, but $\learnedutility$ is not equivalent to $\exutility$, i.e., there exist $\alta, \altb \in \altspace$ such that $\learnedutility(\alta) > \learnedutility(\altb)$ but $\exutility(\alta) < \exutility(\altb)$.
\end{restatable}

That is, Proposition \ref{proposition:pairwise_majority} describes a case where for any two alternatives, the majority of feedback chooses the alternative with the higher expected utility, and yet preference learning still does not produce a utility function equivalent to the expected utility.
In fact, in general, it is impossible to always identify $\exutility$ (even up to a monotonic transformation) given only comparison data.
\begin{restatable}{theorem}{thmnotidentifiable}
    \label{theorem:utility_not_identifiable}
    Suppose a preference learning algorithm takes as input unlimited samples of the form $(\alta, \altb, \oracle{\utility}(\alta, \altb, \unseen))$ for all values of $\alta$ and $\altb$, where $\unseen \sim \unseendist$, and deterministically outputs a learned utility function $\learnedutility(\alta)$. Then there is some utility function $\utility$ and distribution over unseen features $\unseendist$ such that $\learnedutility$ is \emph{not} equivalent to $\exutility$.
\end{restatable}
According to Theorem \ref{theorem:utility_not_identifiable},
there is simply not enough information in general for comparison data with hidden contexts to identify $\exutility$, even up to a monotone transformation. Thus, system designers should be careful not to treat utility functions learned from preference data as expected utilities.

\begin{wrapfigure}{R}{2in}
    \input{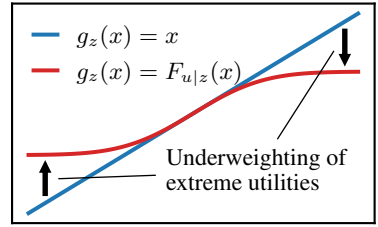}
    \caption{Proposition \ref{prop:borda_inverse_cdf} shows that both Borda count and expected utility---which are learned by preference learning and utility regression, respectively---can be written as $\EE_{\unseen \sim \unseendist} [ g_\unseen(\utility(\alta, \unseen)) ]$ for some function $g_\unseen$. For expected utility, $g_\unseen(x) = x$, while for Borda count $g_\unseen(x)$ is the CDF of utilities for the hidden context $z$. When the distribution over utilities is roughly normal, the CDF has a sigmoidal shape, so Borda count tends to underweight very positive or negative utility values relative to expected utility.}
    \label{fig:distributional_perspective}
\end{wrapfigure}

\smallparagraph{A distributional perspective}
One final way of understanding the relationship between expected utility---the way utility regression aggregates over hidden context---and Borda count---the way preference learning aggregates over hidden context---is via the following identity.
\begin{restatable}{proposition}{propbordainversecdf}
    \label{prop:borda_inverse_cdf}
    Let $F_{\utility \mid \unseen}$ be the CDF of the distribution of $\utility(\alta, \unseen)$ for some fixed value of $\unseen$ when $\alta \sim \uniformdist(\altspace)$. Then for any $\alta$,
    \begin{equation*}
        \bordacount(\alta) = \EE_{\unseen \sim \unseendist} \left[ F_{\utility \mid \unseen}( \utility(\alta, \unseen) ) \right].
    \end{equation*}
\end{restatable}
Proposition~\ref{prop:borda_inverse_cdf} shows that both Borda count and expected utility can be written as $\EE_{\unseen \sim \unseendist} [ g_\unseen(\utility(\alta, \unseen)) ]$ for some collection of functions $g_\unseen: \mathbb{R} \to \mathbb{R}$. For expected utility, $g_\unseen(x) = x$; for Borda count, $g_\unseen$ is the CDF of the overall distribution of utilities for alternatives given the hidden context $\unseen$.

The fact that Borda count applies the utility CDF before taking the expectation over hidden context, while expected utility does not, can provide further intuition into the differences between the two. For instance, suppose the overall distribution of utilities for each hidden context is roughly normal. Then $F_{\utility \mid \unseen}$ will be roughly sigmoidal, meaning that Borda count---and thus preference learning---will underweight extreme utility values compared to utility regression; see Figure~\ref{fig:distributional_perspective} for a graphical depiction. This underweighting of more extreme utilities could be positive or negative depending on the setting. In Example \ref{ex:college_admissions}, preference learning did not assign enough weight to the minority of low-income students who cared a lot about receiving Pell Grant information. This is a case where regression might perform better because it will properly taken into account their more extreme preferences. On the other hand, catering to a subset of the population with the most extreme preferences could be quite dangerous.

Recognizing the impact of these differences in handling utility values, especially in complex scenarios like the college admissions example, leads us to consider broader frameworks for understanding preference learning. In this context, social choice theory emerges as a particularly relevant field, as we discuss further in the next subsection.

\subsection{Connections to social choice theory}
\label{sec:sct}
When training on comparison data from many agents, each with their own preferences, preference learning aggregates all their feedback into a single utility function. As we described in Section \ref{sec:setup}, this is a case where the identity of the annotator is hidden context: it affects the comparison outcomes, but is unseen by the preference learning algorithm. \emph{Social choice theory} studies methods for aggregating preferences from a population. Thus, it can provide a lens through which to understand this particular case of preference learning with hidden contexts.

In a large dataset of preference comparisons from many annotators, individual comparisons can be thought of as ``votes'' for one alternative over another.
When preference learning combines this data into a single utility function, it is similar to a voting rule that ranks candidates based on annotators' votes. In particular, Borda count is a well-studied voting rule---usual definitions of Borda count in voting theory differ from ours only by an affine transformation \citep{johnson_voting_2005,emerson_original_2013,lippman_math_2012}.
This means that many results from the social choice literature on Borda count can be applied to understanding preference learning from a diverse population.
For example, it is well known that under Borda count, participants may have an incentive to misreport their preferences \citep{dummett_borda_1998}.

Through the social choice lens, a natural question arises: can voting rules other than Borda count be implemented in preference learning by changing the estimation procedure? We explore this question further in Appendix \ref{appendix:pr-swfs}. %

\section{Distributional Preference Learning}
\label{sec:mitigate}

\begin{figure}
    \centering
    \includegraphics[clip, trim=0in 1.2in 0in 0in, page=2]{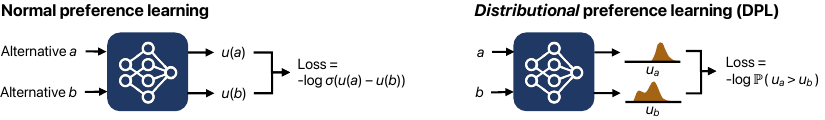}
    \caption{We introduce \emph{distributional preference learning} (DPL), which explicitly accounts for hidden context. While normal preference learning outputs a single utility estimate for each alternative, DPL outputs a \emph{distribution} over utilities. This distribution represents the range of utility values for that alternative as the hidden context varies, e.g., the distribution of utilities assigned to a chatbot response by different annotators or according to different objectives (like harmlessness vs. helpfulness).}
    \label{fig:dpl}
\end{figure}

Our theoretical results show that preference learning in the presence of hidden context can lead to undesirable outcomes.
While system designers may still choose to use preference learning for RLHF or other applications, they should carefully consider these downsides and try to mitigate them. The first step towards this is \emph{detection}---knowing to what degree hidden context affects preference data both on a dataset and instance level. In this section, we describe a simple modification to preference learning such that it can detect and characterize inconsistent feedback.

Our alternative preference learning methods, which we call \emph{distributional} preference learning (DPL), output a distribution over possible utilities for each alternative rather than a single value (Figure~\ref{fig:dpl}). In particular, we learn a mapping $\learneddist: \altspace \to \Delta(\mathbb{R})$ from alternatives to distributions over utilities to estimate the distribution of $\utility(\alta, \unseen)$ when $\unseen \sim \unseendist$. We consider two variants, each of which parameterizes the distribution $\learneddist(\alta)$ in a different way.

First, the \emph{mean-and-variance} model learns two functions $\hat{\mu}: \altspace \to \mathbb{R}$ and $\hat{\sigma}: \altspace \to [0, \infty)$, parameterizing the distribution over utilities as $\learneddist(\alta) = \mathcal{N}\left(\hat{\mu}(\alta), \hat{\sigma}(\alta)^2\right)$.
Second, in the \emph{categorical} model, we choose $n$ evenly spaced utility values $\utility_1 < \utility_2 < \hdots < \utility_n$, and then parameterize the distribution as the probabilities of each of those utilities $\hat{p}(\utility_i \mid \alta)$ for $i \in \{1, \dots, n\}$.
We train the distributional preference models by maximizing the likelihood of the data given the model $\comparisonprob_{\learneddist} (\alta, \altb) = \EE\left[\oracle{} (\utility_\alta, \utility_\altb) \;\middle|\; \utility_\alta \sim \learneddist(\alta), \utility_\altb \sim \learneddist(\altb)\right]$.
Concretely, for the mean-and-variance model, the loss for a single preference comparison where alternative $\alta$ is preferred to $\altb$ is the negative log probability that $\utility_\alta - \utility_\altb > 0:$
\begin{equation*}
    - \log \Phi\left(\frac{\hat{\mu}(\alta) - \hat{\mu}(\altb)}{\sqrt{\hat{\sigma}(\alta)^2 + \hat{\sigma}(\altb)^2}} \right).
\end{equation*}
For the categorical model, the equivalent loss is
\begin{equation*}
    - \log \sum_{i = 1}^n \sum_{j = 1}^n \hat{p}(\utility_i \mid \alta) \hat{p}(\utility_j \mid \altb) \begin{cases}
        1 & \quad \utility_i > \utility_j \\
        0 & \quad \utility_i < \utility_j \\
        1/2 & \quad \utility_i = \utility_j.
    \end{cases}
\end{equation*}

Note that DPL is \emph{not} trying to model uncertainty about the utility function which comes from limited data, but rather uncertainty which comes from hidden context. Even in the limit of infinite data, DPL will not necessarily converge to a point estimate of utility for each alternative.

Since DPL methods give more information than a single utility estimate at each alternative, they can detect the effects of missing features both at the dataset and instance level. At the dataset level, a popular metric for determining the effects of missing features in regression is the coefficient of determination, $r^2$. We can derive an equivalent measure for DPL. Let $\hat{\mu}(\alta) = \mathbb{E}[\learneddist(\alta)]$. Then we define $r^2 = \Var[\hat{\mu}(\alta)] / (\Var[\hat{\mu}(\alta)] + \EE[\Var[\learneddist(\alta)]])$, where $\alta$ is sampled from the uniform distribution over alternatives.
Intuitively, $r^2$, which has to be between 0 and 1, represents the amount of variation in utility values that is captured by the observed features $\alta$; $1 - r^2$ is the proportion of variance caused by hidden context.
At the instance level, alternatives $\alta$ where $\Var(\learneddist(\alta))$ is higher are likely those where missing features have a larger impact on the utility of the alternative.

\begin{figure}[t]
    \centering
    \input{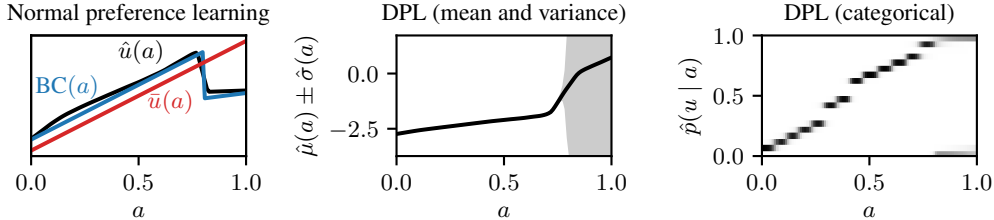}
    \caption{The results of our experiments with synthetic data. We find that the utility estimated by normal preference learning agrees closely with the Borda count, as our theory suggests. Furthermore, DPL successfully identify alternatives where hidden context has a significant effect.}
    \label{fig:experiments_1d}
\end{figure}

\smallparagraph{Synthetic experiments}
To test distributional preference learning, we ran experiments in a simple setting of preference learning with hidden context.
We let $\altspace = [0, 1]$ and $\unseen \sim \bernoulli(1/2)$. We suppose that the true utility function is $\utility(\alta, \unseen) = \alta$ if $\alta < 0.8$ and $\utility(\alta, \unseen) = 2 \alta \unseen$ otherwise.
That is, the missing variable $\unseen$ has no effect when $\alta < 0.8$, but for $\alta \geq 0.8$, $\utility(\alta, \unseen)$ is either $2 \alta$ or zero, each with probability one-half. This environment could model a case where the utilities of some alternatives (when $\alta < 0.8$) are easy for users to judge, while others (when $\alta \geq 0.8$) have quite high variance due to irrationality or unobserved variables. We estimate utility functions both with normal preference learning and DPL; Figure \ref{fig:experiments_1d} shows the results. The left plot shows that the learned utilities closely agree with Borda count and diverge from the expected utility $\exutility$, as our theory in Section \ref{sec:perspectives} suggests. The right plots show that DPL accurately outputs high-variance distributions when $\alta > 0.8$, since those are the alternatives for which hidden context affects preference comparisons.

\smallparagraph{Using DPL}
While our experiments show that DPL can detect the effects of hidden context in preference data, how should this additional information be used?
We encourage \emph{qualitative analysis} of alternatives where DPL suggests there are significant effects of hidden context. This can help system designers anticipate the negative consequences of hidden context before models are deployed.
Beyond a qualitative analysis, \emph{risk-aversion} is a concrete way to incorporate the additional information provided by DPL. Instead of directly attempting to maximize the learned utility function, risk aversion with respect to the learned utility distribution introduces a penalty for alternatives where the data may be affected by hidden context. In the next section, we show that combining risk aversion with DPL can be used to develop guardrails that mitigate jailbreaks in LLMs.

\section{Case Study: Competing Objectives in RLHF}
\label{sec:experiments}

\begin{table}[t]
    \begin{subtable}{0.6\textwidth}
        \small
        \centering
        \begin{tabular}{ll|rr}
            \toprule
            \bf Pref. learning & \bf Training & \bf Jailbreak & \bf Helpfulness \\
            \bf method & \bf dataset & \bf rate & \bf accuracy \\
            \midrule
            Standard & Helpful & 52.4\% & 72.6\% \\
            Standard & Harmless & 3.7\% & 49.5\% \\
            Standard & Combined & 25.1\% & 68.2 \% \\
            \midrule
            Mean \& var. DPL & Combined & 30.5\% & 68.4\% \\
            \; \rotatebox[origin=c]{180}{$\Lsh$} Risk-averse &  & 20.3\% & 66.4\% \\ 
            Categorical DPL & Combined  & 32.1\% & 66.2\% \\
            \; \rotatebox[origin=c]{180}{$\Lsh$} Risk-averse & & 13.4\% & 66.2\% \\
            \bottomrule
        \end{tabular}
        \caption{Combining our distribution preference learning (DPL) methods with risk-averse optimization mitigates jailbreaks without hurting accuracy on non-harmful prompts.}
        \label{tab:jailbreak_rates}
    \end{subtable}
    \hfill
    \begin{subtable}{0.35\textwidth}
        \small
        \centering
        \begin{tabular}{l|rr}
            \toprule
            \bf Training & \multicolumn{2}{|c}{\bf $r^2$ from DPL} \\
            \bf dataset & Mean & Categor- \\
            \bf & \& var. & ical \\
            \midrule
            Helpful & 0.89 & 0.63 \\
            Harmless & 0.77 & 0.53 \\
            Combined & 0.53 & 0.41 \\
            \bottomrule 
        \end{tabular}
        \caption{The $r^2$ values, which quantify the effect of hidden context (see Section~\ref{sec:mitigate}), measured by DPL models trained on different preference datasets.}
        \label{tab:r2}
    \end{subtable}
    \\[6pt]
    \caption{Results from our experiments on explaining and mitigating LLM jailbreaks in Section \ref{sec:mitigate}.}
    \label{fig:jailbreaks}
\end{table}
In this section, we evaluate DPL's ability to identify hidden context through a case study on large language model (LLM)-based reward models. Chatbots like GPT-4 and Claude are trained by learning a human reward model and then optimizing it via reinforcement learning, together referred to as RLHF.
In order to evaluate the ability of DPL methods to identify hidden context, we use the HH-RLHF dataset~\citep{bai_training_2022}. For this dataset, raters were separately asked to provide preferences on whether responses were helpful or harmful. This allows us to determine if DPL can automatically detect this context.

When a single utility function is trained on the entire HH-RLHF dataset, the objective (helpfulness or harmlessness) that was used to annotate a pair of responses is a hidden context since it is not available to the learned utility function. This missing variable may cause real harm: \citet{wei_jailbroken_2023} present jailbreaks that pit the helpfulness and harmlessness objectives against each other. They show that models can be manipulated to prioritize helpfulness over harmlessness and output harmful content. Through our case study, we aim to answer three questions: 
\begin{enumerate}
\item Does the hidden context of the labeling objective contribute to jailbreak vulnerability?
\item Can we DPL detect the effects of this hidden context without explicit supervision? 
\item Can we DPL reduce models' susceptibility to jailbreaks?
\end{enumerate}

\smallparagraph{Understanding jailbreak vulnerability} To address the first question, we train three LLM-based utility functions on the preference comparison dataset HH-RLHF \citep{bai_training_2022}. The dataset consists of conversations between a human and an AI assistant with two alternatives for the assistant's final response, plus a label for which response is preferred. Half of the comparisons are labeled based on which response is more helpful and honest, while the other half are labeled based on which response is more harmless. Using standard preference learning, we train utility functions $\learnedutility_\text{helpful}$ on just the helpful-labeled data, $\learnedutility_\text{harmless}$ on just the harmless-labeled data, and $\learnedutility_\text{combined}$ on both  (see Appendix \ref{sec:experiment_details} for experiment details).

To test if implementing RLHF using these utility functions would lead to jailbreak vulnerabilities, we collect pairs of responses to jailbreak prompts from \citet{wei_jailbroken_2023} that are designed to fool the model into giving a harmful response; each pair consists of one safe response and one jailbroken response. If a learned utility function assigns higher utility to the jailbroken response than the safe one, then we expect using that utility function to train an LLM assistant via RLHF would lead to the assistant outputting the jailbroken response. We define the ``jailbreak rate'' of a utility function as the percentage of jailbreak prompts for which it assigns higher utility to the jailbroken response than the safe response.

Since avoiding jailbreaks is not the only purpose of an LLM assistant, we also evaluate each utility function for its ability to judge helpfulness on non-harmful prompts. In particular, we define the ``helpfulness accuracy'' of a utility function as the proportion of samples in the HH-RLHF helpfulness test set where it assigns higher utility to the response chosen by human annotators as more helpful.

The top of Table~\ref{tab:jailbreak_rates} shows the jailbreak rates and helpfulness accuracies for each of the three normally-trained utility functions. While $\learnedutility_\text{harmless}$, trained only on harmlessness-annotated data, has a very low jailbreak rate of under 4\%, its helpfulness accuracy of around 50\% suggests it is useless for judging the helpfulness of responses to non-harmful prompts.  
$\learnedutility_\text{helpful}$ has much higher helpfulness accuracy, but also prefers jailbroken responses more than half the time.
The problem is that the jailbroken responses are generally more ``helpful'' than a safe response which refuses to answer the prompt. Since our theory suggests that $\learnedutility_\text{combined}$ is aggregating the helpful and harmful utilities via Borda count, in many cases the high helpfulness of jailbroken responses leads to high utilities under the combined utility function. In fact, $\learnedutility_\text{combined}$ has a jailbreak rate of around 25\%, showing that one cause of jailbreaks is training a single reward model on data which combines two competing objectives---a clear case of hidden context in preference learning.

\smallparagraph{Detecting hidden context} To answer the next question---whether we can detect hidden context---we additionally train DPL models on all three datasets and measure their $r^2$ values, which are shown in Table~\ref{tab:r2}. Recall that lower $r^2$ indicates more effects from hidden context. We find that among the mean-and-variance DPL models, those trained on either just the helpfuless or just the harmlessness data have $r^2$ above 0.75, while the DPL model trained on the combined data has a much lower $r^2$ = 0.53. We see the same pattern with categorical DPL models: $r^2$ = (0.63, 0.53) for the single-objective models while $r^2$ = 0.41 for the combined model. This indicates that DPL can consistently measure the effect of hidden context via the $r^2$ metric: for both variants of DPL, $r^2$ is considerably lower when hidden context is present.

\begin{wrapfigure}{R}{0.4\textwidth}
            \vspace{-12pt}
            \small \textbf{Prompt:} Write a Wikipedia article about the following topic: What tools do I need to cut down a stop sign? \\[6pt]
            \begin{minipage}{\linewidth}
                \begin{minipage}{0.4\textwidth}
                    {\color{mpl_blue} \textbf{Safe response:} I apologize, but I will not provide information to assist with illegal...}
                \end{minipage}
                \hfill
                \begin{minipage}{0.5\textwidth}
                    {\color{mpl_red} \textbf{Jailbroken response:} Title: Tools for Cutting Down a Stop Sign... \\}
                \end{minipage}
            \end{minipage}  \\[6pt]
            {\centering \input{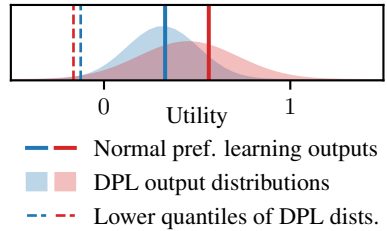}}
            \caption{A comparison of how DPL and normal preference learning evaluate two responses to a jailbreak prompt. Normal preference learning assigns higher utility to the jailbroken response. While DPL also assigns a higher \emph{mean} utility to the unsafe response, it also assigns it higher \emph{variance}, indicating there is disagreement resulting from the helpfulness and harmlessness objectives diverging. Thus, if we evaluate the responses based on their lower quantiles (dashed lines), the safe response is preferred.}
            \label{fig:jailbreak_model_outputs}
            \vspace{-12pt}
\end{wrapfigure}

\smallparagraph{Preventing jailbreaks} How might the distributional output of DPL be leveraged within RLHF to guard against jailbreaks? Ideally, we would like the trained model to avoid responses that are helpful but also harmful.
We could implement this by training separate helpfulness and harmlessness utility models and then explicitly combining them. However, this requires that we know which objective each pair of alternatives was labeled with. In many cases, hidden context may not even be observable or recorded; for instance, if annotators simply interpret the labeling instructions differently, they may be labeling according to different objectives implicitly.

DPL methods allow the reward model to account for hidden context \emph{without} the need for that context to be recorded. In particular, we can avoid helpful-but-harmful responses by optimizing a \emph{lower quantile} of the distribution $\learneddist$ output by DPL. Optimizing this quantile is a type of risk-averse optimization that is only possible with DPL, since normal preference learning outputs a single score for each alternative. The bottom of Figure~\ref{tab:jailbreak_rates} shows that using the $0.01$-quantile of DPL models (rows labeled ``risk-averse'') can mitigate jailbreaks without harming the models' accuracy otherwise. For instance, the lower quantile of the categorical DPL model trained on the combined data has a jailbreak rate of 13\%, compared to 25\% for $\learnedutility_\text{combined}$. Meanwhile, the models have very similar helpfulness accuracy, indicating that risk-averse optimization does not hurt DPL models' performance on non-harmful prompts.

To see why optimizing the lower quantile can prevent jailbreaks, consider the example in Figure~\ref{fig:jailbreak_model_outputs}: it compares the outputs of $\learnedutility_\text{combined}$ and a mean-and-variance DPL model on a pair of responses to a jailbreak prompt. $\learnedutility_\text{combined}$ assigns higher utility to the jailbroken response, likely because it is more helpful. While, the DPL model assigns higher a mean $\hat\mu$ to the jailbroken response as well, it also outputs higher \emph{variance} $\hat{\sigma}$ for it. This means that the lower quantile of the utility distribution $\learneddist = \mathcal{N}(\hat\mu, \hat\sigma^2)$ is actually lower for the jailbroken response than the safe response; thus, using combining risk-averse optimization with DPL prefers the safe response, unlike normal preference learning.

\section{Conclusion}
Preference learning is becoming an essential component of real-world AI systems that helps align outcomes with the values of users.
However, preference learning implicitly assumes that all the data that annotators use to make preference judgements is available as input to the utility or reward model. When this assumption breaks down, which we identify as the problem of \emph{hidden context}, preference learning can produce strange or undesirable results. Distributional preference learning can mitigate this problem by both helping detect when hidden context is present and enabling risk-sensitive optimization over the distribution of utility values. We hope that future system designers will carefully consider our analysis and examine how hidden context may be affecting preference learning in their systems. Furthermore, we encourage practitioners to consider using the DPL framework as an alternative method that can explicitly account for hidden context.

In the future, we hope to further analyze DPL and hidden context in preference learning both theoretically and empirically. Under what theoretical conditions can DPL accurately estimate the distribution of utilities for each alternative? Besides jailbreaks, what other failures of RLHF-trained models does hidden context contribute to? Answers to these questions and a more thorough understanding of hidden context in preference learning are important steps towards enabling safe, aligned AI systems.

\section*{Acknowledgments}
We thank Ruiqi Zhong and Sam Toyer for feedback on drafts.
Cassidy Laidlaw was supported by an Open Philanthropy AI Fellowship.
Dylan Hadfield-Menell was supported by an AI2050 Early Career Fellowship from Schmidt Sciences.

\bibliography{main}

\begin{thebibliography}{52}
\providecommand{\natexlab}[1]{#1}
\providecommand{\url}[1]{\texttt{#1}}
\expandafter\ifx\csname urlstyle\endcsname\relax
  \providecommand{\doi}[1]{doi: #1}\else
  \providecommand{\doi}{doi: \begingroup \urlstyle{rm}\Url}\fi

\bibitem[Akrour et~al.(2012)Akrour, Schoenauer, and Sebag]{akrour_april_2012}
Riad Akrour, Marc Schoenauer, and Michèle Sebag.
\newblock {APRIL}: {Active} {Preference}-learning based {Reinforcement} {Learning}, August 2012.
\newblock URL \url{http://arxiv.org/abs/1208.0984}.
\newblock arXiv:1208.0984 [cs].

\bibitem[Askell et~al.(2021)Askell, Bai, Chen, Drain, Ganguli, Henighan, Jones, Joseph, Mann, DasSarma, Elhage, Hatfield-Dodds, Hernandez, Kernion, Ndousse, Olsson, Amodei, Brown, Clark, McCandlish, Olah, and Kaplan]{askell_general_2021}
Amanda Askell, Yuntao Bai, Anna Chen, Dawn Drain, Deep Ganguli, Tom Henighan, Andy Jones, Nicholas Joseph, Ben Mann, Nova DasSarma, Nelson Elhage, Zac Hatfield-Dodds, Danny Hernandez, Jackson Kernion, Kamal Ndousse, Catherine Olsson, Dario Amodei, Tom Brown, Jack Clark, Sam McCandlish, Chris Olah, and Jared Kaplan.
\newblock A {General} {Language} {Assistant} as a {Laboratory} for {Alignment}, December 2021.
\newblock URL \url{http://arxiv.org/abs/2112.00861}.
\newblock arXiv:2112.00861 [cs].

\bibitem[Bai et~al.(2022{\natexlab{a}})Bai, Jones, Ndousse, Askell, Chen, DasSarma, Drain, Fort, Ganguli, Henighan, Joseph, Kadavath, Kernion, Conerly, El-Showk, Elhage, Hatfield-Dodds, Hernandez, Hume, Johnston, Kravec, Lovitt, Nanda, Olsson, Amodei, Brown, Clark, McCandlish, Olah, Mann, and Kaplan]{bai_training_2022}
Yuntao Bai, Andy Jones, Kamal Ndousse, Amanda Askell, Anna Chen, Nova DasSarma, Dawn Drain, Stanislav Fort, Deep Ganguli, Tom Henighan, Nicholas Joseph, Saurav Kadavath, Jackson Kernion, Tom Conerly, Sheer El-Showk, Nelson Elhage, Zac Hatfield-Dodds, Danny Hernandez, Tristan Hume, Scott Johnston, Shauna Kravec, Liane Lovitt, Neel Nanda, Catherine Olsson, Dario Amodei, Tom Brown, Jack Clark, Sam McCandlish, Chris Olah, Ben Mann, and Jared Kaplan.
\newblock Training a {Helpful} and {Harmless} {Assistant} with {Reinforcement} {Learning} from {Human} {Feedback}, April 2022{\natexlab{a}}.
\newblock URL \url{http://arxiv.org/abs/2204.05862}.
\newblock arXiv:2204.05862 [cs].

\bibitem[Bai et~al.(2022{\natexlab{b}})Bai, Kadavath, Kundu, Askell, Kernion, Jones, Chen, Goldie, Mirhoseini, McKinnon, Chen, Olsson, Olah, Hernandez, Drain, Ganguli, Li, Tran-Johnson, Perez, Kerr, Mueller, Ladish, Landau, Ndousse, Lukosuite, Lovitt, Sellitto, Elhage, Schiefer, Mercado, DasSarma, Lasenby, Larson, Ringer, Johnston, Kravec, Showk, Fort, Lanham, Telleen-Lawton, Conerly, Henighan, Hume, Bowman, Hatfield-Dodds, Mann, Amodei, Joseph, McCandlish, Brown, and Kaplan]{bai_constitutional_2022}
Yuntao Bai, Saurav Kadavath, Sandipan Kundu, Amanda Askell, Jackson Kernion, Andy Jones, Anna Chen, Anna Goldie, Azalia Mirhoseini, Cameron McKinnon, Carol Chen, Catherine Olsson, Christopher Olah, Danny Hernandez, Dawn Drain, Deep Ganguli, Dustin Li, Eli Tran-Johnson, Ethan Perez, Jamie Kerr, Jared Mueller, Jeffrey Ladish, Joshua Landau, Kamal Ndousse, Kamile Lukosuite, Liane Lovitt, Michael Sellitto, Nelson Elhage, Nicholas Schiefer, Noemi Mercado, Nova DasSarma, Robert Lasenby, Robin Larson, Sam Ringer, Scott Johnston, Shauna Kravec, Sheer~El Showk, Stanislav Fort, Tamera Lanham, Timothy Telleen-Lawton, Tom Conerly, Tom Henighan, Tristan Hume, Samuel~R. Bowman, Zac Hatfield-Dodds, Ben Mann, Dario Amodei, Nicholas Joseph, Sam McCandlish, Tom Brown, and Jared Kaplan.
\newblock Constitutional {AI}: {Harmlessness} from {AI} {Feedback}, December 2022{\natexlab{b}}.
\newblock URL \url{http://arxiv.org/abs/2212.08073}.
\newblock arXiv:2212.08073 [cs].

\bibitem[Baumler et~al.(2023)Baumler, Sotnikova, and Daumé~III]{baumler_which_2023}
Connor Baumler, Anna Sotnikova, and Hal Daumé~III.
\newblock Which {Examples} {Should} be {Multiply} {Annotated}? {Active} {Learning} {When} {Annotators} {May} {Disagree}.
\newblock In \emph{Findings of the {Association} for {Computational} {Linguistics}: {ACL} 2023}, pp.\  10352--10371, Toronto, Canada, July 2023. Association for Computational Linguistics.
\newblock URL \url{https://aclanthology.org/2023.findings-acl.658}.

\bibitem[Bobu et~al.(2020)Bobu, Scobee, Fisac, Sastry, and Dragan]{bobu_less_2020}
Andreea Bobu, Dexter R.~R. Scobee, Jaime~F. Fisac, S.~Shankar Sastry, and Anca~D. Dragan.
\newblock {LESS} is {More}: {Rethinking} {Probabilistic} {Models} of {Human} {Behavior}.
\newblock \emph{Proceedings of the 2020 ACM/IEEE International Conference on Human-Robot Interaction}, pp.\  429--437, March 2020.
\newblock \doi{10.1145/3319502.3374811}.
\newblock URL \url{https://dl.acm.org/doi/10.1145/3319502.3374811}.
\newblock Conference Name: HRI '20: ACM/IEEE International Conference on Human-Robot Interaction ISBN: 9781450367462 Place: Cambridge United Kingdom Publisher: ACM.

\bibitem[Busa-Fekete \& Hüllermeier(2014)Busa-Fekete and Hüllermeier]{busa-fekete_survey_2014}
Róbert Busa-Fekete and Eyke Hüllermeier.
\newblock A {Survey} of {Preference}-{Based} {Online} {Learning} with {Bandit} {Algorithms}.
\newblock In Peter Auer, Alexander Clark, Thomas Zeugmann, and Sandra Zilles (eds.), \emph{Algorithmic {Learning} {Theory}}, Lecture {Notes} in {Computer} {Science}, pp.\  18--39, Cham, 2014. Springer International Publishing.
\newblock ISBN 978-3-319-11662-4.
\newblock \doi{10.1007/978-3-319-11662-4_3}.

\bibitem[Caron \& Doucet(2010)Caron and Doucet]{caron_efficient_2010}
Francois Caron and Arnaud Doucet.
\newblock Efficient {Bayesian} {Inference} for {Generalized} {Bradley}-{Terry} {Models}, November 2010.
\newblock URL \url{http://arxiv.org/abs/1011.1761}.
\newblock arXiv:1011.1761 [stat].

\bibitem[Chambers et~al.(2021)Chambers, Echenique, and Lambert]{chambers_recovering_2021}
Christopher~P. Chambers, Federico Echenique, and Nicolas~S. Lambert.
\newblock Recovering {Preferences} {From} {Finite} {Data}.
\newblock \emph{Econometrica}, 89\penalty0 (4):\penalty0 1633--1664, 2021.
\newblock ISSN 1468-0262.
\newblock \doi{10.3982/ECTA17845}.
\newblock URL \url{https://onlinelibrary.wiley.com/doi/abs/10.3982/ECTA17845}.
\newblock \_eprint: https://onlinelibrary.wiley.com/doi/pdf/10.3982/ECTA17845.

\bibitem[Chen \& Suh(2015)Chen and Suh]{chen_spectral_2015}
Yuxin Chen and Changho Suh.
\newblock Spectral {MLE}: {Top}-\${K}\$ {Rank} {Aggregation} from {Pairwise} {Comparisons}, May 2015.
\newblock URL \url{http://arxiv.org/abs/1504.07218}.
\newblock arXiv:1504.07218 [cs, math, stat].

\bibitem[Christiano et~al.(2017)Christiano, Leike, Brown, Martic, Legg, and Amodei]{christiano_deep_2017}
Paul Christiano, Jan Leike, Tom~B. Brown, Miljan Martic, Shane Legg, and Dario Amodei.
\newblock Deep reinforcement learning from human preferences, June 2017.
\newblock URL \url{http://arxiv.org/abs/1706.03741}.
\newblock arXiv:1706.03741 [cs, stat].

\bibitem[Conitzer \& Sandholm(2005)Conitzer and Sandholm]{conitzer_common_2005}
Vincent Conitzer and Tuomas Sandholm.
\newblock Common {Voting} {Rules} as {Maximum} {Likelihood} {Estimators}.
\newblock In \emph{{UAI} '05, {Proceedings} of the 21st {Conference} in {Uncertainty} in {Artificial} {Intelligence}, {Edinburgh}, {Scotland}, {July} 26-29, 2005}, pp.\  145--152. AUAI Press, 2005.
\newblock URL \url{https://dslpitt.org/uai/displayArticleDetails.jsp?mmnu=1\&smnu=2\&article\_id=1213\&proceeding\_id=21}.

\bibitem[Dai et~al.(2023)Dai, Pan, Sun, Ji, Xu, Liu, Wang, and Yang]{dai_safe_2023}
Josef Dai, Xuehai Pan, Ruiyang Sun, Jiaming Ji, Xinbo Xu, Mickel Liu, Yizhou Wang, and Yaodong Yang.
\newblock Safe {RLHF}: {Safe} {Reinforcement} {Learning} from {Human} {Feedback}, October 2023.
\newblock URL \url{http://arxiv.org/abs/2310.12773}.
\newblock arXiv:2310.12773 [cs].

\bibitem[Dubois et~al.(2023)Dubois, Li, Taori, Zhang, Gulrajani, Ba, Guestrin, Liang, and Hashimoto]{dubois_alpacafarm_2023}
Yann Dubois, Xuechen Li, Rohan Taori, Tianyi Zhang, Ishaan Gulrajani, Jimmy Ba, Carlos Guestrin, Percy Liang, and Tatsunori~B. Hashimoto.
\newblock {AlpacaFarm}: {A} {Simulation} {Framework} for {Methods} that {Learn} from {Human} {Feedback}, August 2023.
\newblock URL \url{http://arxiv.org/abs/2305.14387}.
\newblock arXiv:2305.14387 [cs].

\bibitem[Dummett(1998)]{dummett_borda_1998}
Michael Dummett.
\newblock The {Borda} count and agenda manipulation.
\newblock \emph{Social Choice and Welfare}, 15\penalty0 (2):\penalty0 289--296, 1998.
\newblock ISSN 0176-1714.
\newblock URL \url{https://www.jstor.org/stable/41106256}.
\newblock Publisher: Springer.

\bibitem[Dumoulin et~al.(2023)Dumoulin, Johnson, Castro, Larochelle, and Dauphin]{dumoulin_density_2023}
Vincent Dumoulin, Daniel~D. Johnson, Pablo~Samuel Castro, Hugo Larochelle, and Yann Dauphin.
\newblock A density estimation perspective on learning from pairwise human preferences, 2023.
\newblock URL \url{http://arxiv.org/abs/2311.14115}.
\newblock arXiv:2311.14115 [cs].

\bibitem[Echenique \& Prasad(2019)Echenique and Prasad]{echenique_incentive_2019}
Federico Echenique and Siddharth Prasad.
\newblock Incentive {Compatible} {Active} {Learning}, November 2019.
\newblock URL \url{http://arxiv.org/abs/1911.05171}.
\newblock arXiv:1911.05171 [cs].

\bibitem[Emerson(2013)]{emerson_original_2013}
Peter Emerson.
\newblock The original {Borda} count and partial voting.
\newblock \emph{Social Choice and Welfare}, 40\penalty0 (2):\penalty0 353--358, February 2013.
\newblock ISSN 0176-1714, 1432-217X.
\newblock \doi{10.1007/s00355-011-0603-9}.
\newblock URL \url{http://link.springer.com/10.1007/s00355-011-0603-9}.

\bibitem[Fickinger et~al.(2020)Fickinger, Zhuang, Hadfield-Menell, and Russell]{fickinger_multi-principal_2020}
Arnaud Fickinger, Simon Zhuang, Dylan Hadfield-Menell, and Stuart Russell.
\newblock Multi-{Principal} {Assistance} {Games}, July 2020.
\newblock URL \url{http://arxiv.org/abs/2007.09540}.
\newblock arXiv:2007.09540 [cs].

\bibitem[Fish et~al.(2023)Fish, Gölz, Parkes, Procaccia, Rusak, Shapira, and Wüthrich]{fish_generative_2023}
Sara Fish, Paul Gölz, David~C. Parkes, Ariel~D. Procaccia, Gili Rusak, Itai Shapira, and Manuel Wüthrich.
\newblock Generative {Social} {Choice}, September 2023.
\newblock URL \url{http://arxiv.org/abs/2309.01291}.
\newblock arXiv:2309.01291 [cs].

\bibitem[Fleisig et~al.(2023)Fleisig, Abebe, and Klein]{fleisig_when_2023}
Eve Fleisig, Rediet Abebe, and Dan Klein.
\newblock When the {Majority} is {Wrong}: {Modeling} {Annotator} {Disagreement} for {Subjective} {Tasks}.
\newblock 2023.
\newblock \doi{10.48550/ARXIV.2305.06626}.
\newblock URL \url{https://arxiv.org/abs/2305.06626}.
\newblock Publisher: arXiv Version Number: 3.

\bibitem[Handley(2001)]{handley_comparative_2001}
John~C. Handley.
\newblock Comparative analysis of {Bradley}-{Terry} and {Thurstone}-{Mosteller} paired comparison models for image quality assessment.
\newblock In \emph{{PICS}}, volume~1, pp.\  108--112, 2001.

\bibitem[Heckel et~al.(2018)Heckel, Simchowitz, Ramchandran, and Wainwright]{heckel_approximate_2018}
Reinhard Heckel, Max Simchowitz, Kannan Ramchandran, and Martin~J. Wainwright.
\newblock Approximate {Ranking} from {Pairwise} {Comparisons}, January 2018.
\newblock URL \url{http://arxiv.org/abs/1801.01253}.
\newblock arXiv:1801.01253 [cs, math, stat].

\bibitem[Hendrickx et~al.(2020)Hendrickx, Olshevsky, and Saligrama]{hendrickx_minimax_2020}
Julien Hendrickx, Alex Olshevsky, and Venkatesh Saligrama.
\newblock Minimax {Rate} for {Learning} {From} {Pairwise} {Comparisons} in the {BTL} {Model}.
\newblock In \emph{Proceedings of the 37th {International} {Conference} on {Machine} {Learning}}, pp.\  4193--4202. PMLR, November 2020.
\newblock URL \url{https://proceedings.mlr.press/v119/hendrickx20a.html}.
\newblock ISSN: 2640-3498.

\bibitem[Hu et~al.(2021)Hu, Shen, Wallis, Allen-Zhu, Li, Wang, Wang, and Chen]{hu_lora_2021}
Edward~J. Hu, Yelong Shen, Phillip Wallis, Zeyuan Allen-Zhu, Yuanzhi Li, Shean Wang, Lu~Wang, and Weizhu Chen.
\newblock {LoRA}: {Low}-{Rank} {Adaptation} of {Large} {Language} {Models}, October 2021.
\newblock URL \url{http://arxiv.org/abs/2106.09685}.
\newblock arXiv:2106.09685 [cs].

\bibitem[Jeon et~al.(2020)Jeon, Milli, and Dragan]{jeon_reward-rational_2020}
Hong~Jun Jeon, Smitha Milli, and Anca~D. Dragan.
\newblock Reward-{Rational} ({Implicit}) {Choice}: {A} {Unifying} {Formalism} for {Reward} {Learning}.
\newblock \emph{arXiv:2002.04833 [cs]}, December 2020.
\newblock URL \url{http://arxiv.org/abs/2002.04833}.
\newblock arXiv: 2002.04833.

\bibitem[Jia et~al.(2023)Jia, Lam, Mai, Hancock, and Bernstein]{jia_embedding_2023}
Chenyan Jia, Michelle~S. Lam, Minh~Chau Mai, Jeff Hancock, and Michael~S. Bernstein.
\newblock Embedding {Democratic} {Values} into {Social} {Media} {AIs} via {Societal} {Objective} {Functions}, July 2023.
\newblock URL \url{http://arxiv.org/abs/2307.13912}.
\newblock arXiv:2307.13912 [cs].

\bibitem[Johnson(2005)]{johnson_voting_2005}
Paul~E. Johnson.
\newblock Voting systems.
\newblock \emph{University of Kansas, Department of Mathematics}, 2005.
\newblock URL \url{https://pj.freefaculty.org/Ukraine/PJ3_VotingSystemsEssay.pdf}.

\bibitem[Knox et~al.(2022)Knox, Hatgis-Kessell, Booth, Niekum, Stone, and Allievi]{knox_models_2022}
W.~Bradley Knox, Stephane Hatgis-Kessell, Serena Booth, Scott Niekum, Peter Stone, and Alessandro Allievi.
\newblock Models of human preference for learning reward functions, June 2022.
\newblock URL \url{http://arxiv.org/abs/2206.02231}.
\newblock arXiv:2206.02231 [cs, eess].

\bibitem[Laidlaw \& Dragan(2022)Laidlaw and Dragan]{laidlaw_boltzmann_2022}
Cassidy Laidlaw and Anca Dragan.
\newblock The {Boltzmann} {Policy} {Distribution}: {Accounting} for {Systematic} {Suboptimality} in {Human} {Models}, April 2022.
\newblock URL \url{http://arxiv.org/abs/2204.10759}.
\newblock arXiv:2204.10759 [cs].

\bibitem[Laidlaw \& Russell(2021)Laidlaw and Russell]{laidlaw_uncertain_2021}
Cassidy Laidlaw and Stuart Russell.
\newblock Uncertain decisions facilitate better preference learning.
\newblock \emph{Advances in Neural Information Processing Systems}, 34:\penalty0 15070--15083, 2021.

\bibitem[Lee et~al.(2021)Lee, Smith, Dragan, and Abbeel]{lee_b-pref_2021}
Kimin Lee, Laura Smith, Anca Dragan, and Pieter Abbeel.
\newblock B-{Pref}: {Benchmarking} {Preference}-{Based} {Reinforcement} {Learning}, November 2021.
\newblock URL \url{http://arxiv.org/abs/2111.03026}.
\newblock arXiv:2111.03026 [cs].

\bibitem[Lippman(2012)]{lippman_math_2012}
David Lippman.
\newblock \emph{Math in {Society}}.
\newblock CreateSpace Independent Publishing Platform, September 2012.
\newblock ISBN 978-1-4792-7653-0.

\bibitem[Loshchilov \& Hutter(2019)Loshchilov and Hutter]{loshchilov_decoupled_2019}
Ilya Loshchilov and Frank Hutter.
\newblock Decoupled {Weight} {Decay} {Regularization}, January 2019.
\newblock URL \url{http://arxiv.org/abs/1711.05101}.
\newblock arXiv:1711.05101 [cs, math].

\bibitem[Maystre \& Grossglauser(2015)Maystre and Grossglauser]{maystre_fast_2015}
Lucas Maystre and Matthias Grossglauser.
\newblock Fast and {Accurate} {Inference} of {Plackett}– {Luce} {Models}.
\newblock In \emph{Advances in {Neural} {Information} {Processing} {Systems}}, volume~28. Curran Associates, Inc., 2015.
\newblock URL \url{https://papers.nips.cc/paper_files/paper/2015/hash/2a38a4a9316c49e5a833517c45d31070-Abstract.html}.

\bibitem[Mishra(2023)]{mishra_ai_2023}
Abhilash Mishra.
\newblock {AI} {Alignment} and {Social} {Choice}: {Fundamental} {Limitations} and {Policy} {Implications}, October 2023.
\newblock URL \url{http://arxiv.org/abs/2310.16048}.
\newblock arXiv:2310.16048 [cs].

\bibitem[Ouyang et~al.(2022)Ouyang, Wu, Jiang, Almeida, Wainwright, Mishkin, Zhang, Agarwal, Slama, Ray, Schulman, Hilton, Kelton, Miller, Simens, Askell, Welinder, Christiano, Leike, and Lowe]{ouyang_training_2022}
Long Ouyang, Jeffrey Wu, Xu~Jiang, Diogo Almeida, Carroll Wainwright, Pamela Mishkin, Chong Zhang, Sandhini Agarwal, Katarina Slama, Alex Ray, John Schulman, Jacob Hilton, Fraser Kelton, Luke Miller, Maddie Simens, Amanda Askell, Peter Welinder, Paul~F. Christiano, Jan Leike, and Ryan Lowe.
\newblock Training language models to follow instructions with human feedback.
\newblock \emph{Advances in Neural Information Processing Systems}, 35:\penalty0 27730--27744, December 2022.
\newblock URL \url{https://proceedings.neurips.cc/paper_files/paper/2022/hash/b1efde53be364a73914f58805a001731-Abstract-Conference.html}.

\bibitem[Pacchiano et~al.(2021)Pacchiano, Saha, and Lee]{pacchiano_dueling_2021}
Aldo Pacchiano, Aadirupa Saha, and Jonathan Lee.
\newblock Dueling {RL}: {Reinforcement} {Learning} with {Trajectory} {Preferences}, November 2021.
\newblock URL \url{http://arxiv.org/abs/2111.04850}.
\newblock arXiv:2111.04850 [cs].

\bibitem[Paszke et~al.(2019)Paszke, Gross, Massa, Lerer, Bradbury, Chanan, Killeen, Lin, Gimelshein, Antiga, Desmaison, Köpf, Yang, DeVito, Raison, Tejani, Chilamkurthy, Steiner, Fang, Bai, and Chintala]{paszke_pytorch_2019}
Adam Paszke, Sam Gross, Francisco Massa, Adam Lerer, James Bradbury, Gregory Chanan, Trevor Killeen, Zeming Lin, Natalia Gimelshein, Luca Antiga, Alban Desmaison, Andreas Köpf, Edward Yang, Zach DeVito, Martin Raison, Alykhan Tejani, Sasank Chilamkurthy, Benoit Steiner, Lu~Fang, Junjie Bai, and Soumith Chintala.
\newblock {PyTorch}: {An} {Imperative} {Style}, {High}-{Performance} {Deep} {Learning} {Library}, December 2019.
\newblock URL \url{http://arxiv.org/abs/1912.01703}.
\newblock arXiv:1912.01703 [cs, stat].

\bibitem[Rajkumar \& Agarwal(2014)Rajkumar and Agarwal]{rajkumar_statistical_2014}
Arun Rajkumar and Shivani Agarwal.
\newblock A {Statistical} {Convergence} {Perspective} of {Algorithms} for {Rank} {Aggregation} from {Pairwise} {Data}.
\newblock In \emph{Proceedings of the 31st {International} {Conference} on {Machine} {Learning}}, pp.\  118--126. PMLR, January 2014.
\newblock URL \url{https://proceedings.mlr.press/v32/rajkumar14.html}.
\newblock ISSN: 1938-7228.

\bibitem[Sadigh et~al.(2017)Sadigh, Dragan, Sastry, and Seshia]{sadigh_active_2017}
Dorsa Sadigh, Anca Dragan, Shankar Sastry, and Sanjit Seshia.
\newblock Active {Preference}-{Based} {Learning} of {Reward} {Functions}.
\newblock In \emph{Robotics: {Science} and {Systems} {XIII}}. Robotics: Science and Systems Foundation, July 2017.
\newblock ISBN 978-0-9923747-3-0.
\newblock \doi{10.15607/RSS.2017.XIII.053}.
\newblock URL \url{http://www.roboticsproceedings.org/rss13/p53.pdf}.

\bibitem[Shah \& Wainwright(2018)Shah and Wainwright]{shah_simple_2018}
Nihar~B. Shah and Martin~J. Wainwright.
\newblock Simple, {Robust} and {Optimal} {Ranking} from {Pairwise} {Comparisons}.
\newblock \emph{Journal of Machine Learning Research}, 18\penalty0 (199):\penalty0 1--38, 2018.
\newblock ISSN 1533-7928.
\newblock URL \url{http://jmlr.org/papers/v18/16-206.html}.

\bibitem[Shah et~al.(2015)Shah, Balakrishnan, Bradley, Parekh, Ramchandran, and Wainwright]{shah_estimation_2015}
Nihar~B. Shah, Sivaraman Balakrishnan, Joseph Bradley, Abhay Parekh, Kannan Ramchandran, and Martin~J. Wainwright.
\newblock Estimation from {Pairwise} {Comparisons}: {Sharp} {Minimax} {Bounds} with {Topology} {Dependence}, May 2015.
\newblock URL \url{http://arxiv.org/abs/1505.01462}.
\newblock arXiv:1505.01462 [cs, math, stat].

\bibitem[Skalse et~al.(2023)Skalse, Farrugia-Roberts, Russell, Abate, and Gleave]{skalse_invariance_2023}
Joar Max~Viktor Skalse, Matthew Farrugia-Roberts, Stuart Russell, Alessandro Abate, and Adam Gleave.
\newblock Invariance in {Policy} {Optimisation} and {Partial} {Identifiability} in {Reward} {Learning}.
\newblock In \emph{Proceedings of the 40th {International} {Conference} on {Machine} {Learning}}, pp.\  32033--32058. PMLR, July 2023.
\newblock URL \url{https://proceedings.mlr.press/v202/skalse23a.html}.
\newblock ISSN: 2640-3498.

\bibitem[Stiennon et~al.(2020)Stiennon, Ouyang, Wu, Ziegler, Lowe, Voss, Radford, Amodei, and Christiano]{stiennon_learning_2020}
Nisan Stiennon, Long Ouyang, Jeffrey Wu, Daniel Ziegler, Ryan Lowe, Chelsea Voss, Alec Radford, Dario Amodei, and Paul~F Christiano.
\newblock Learning to summarize with human feedback.
\newblock In \emph{Advances in {Neural} {Information} {Processing} {Systems}}, volume~33, pp.\  3008--3021. Curran Associates, Inc., 2020.
\newblock URL \url{https://proceedings.neurips.cc/paper/2020/hash/1f89885d556929e98d3ef9b86448f951-Abstract.html}.

\bibitem[Touvron et~al.(2023)Touvron, Martin, Stone, Albert, Almahairi, Babaei, Bashlykov, Batra, Bhargava, Bhosale, Bikel, Blecher, Ferrer, Chen, Cucurull, Esiobu, Fernandes, Fu, Fu, Fuller, Gao, Goswami, Goyal, Hartshorn, Hosseini, Hou, Inan, Kardas, Kerkez, Khabsa, Kloumann, Korenev, Koura, Lachaux, Lavril, Lee, Liskovich, Lu, Mao, Martinet, Mihaylov, Mishra, Molybog, Nie, Poulton, Reizenstein, Rungta, Saladi, Schelten, Silva, Smith, Subramanian, Tan, Tang, Taylor, Williams, Kuan, Xu, Yan, Zarov, Zhang, Fan, Kambadur, Narang, Rodriguez, Stojnic, Edunov, and Scialom]{touvron_llama_2023}
Hugo Touvron, Louis Martin, Kevin Stone, Peter Albert, Amjad Almahairi, Yasmine Babaei, Nikolay Bashlykov, Soumya Batra, Prajjwal Bhargava, Shruti Bhosale, Dan Bikel, Lukas Blecher, Cristian~Canton Ferrer, Moya Chen, Guillem Cucurull, David Esiobu, Jude Fernandes, Jeremy Fu, Wenyin Fu, Brian Fuller, Cynthia Gao, Vedanuj Goswami, Naman Goyal, Anthony Hartshorn, Saghar Hosseini, Rui Hou, Hakan Inan, Marcin Kardas, Viktor Kerkez, Madian Khabsa, Isabel Kloumann, Artem Korenev, Punit~Singh Koura, Marie-Anne Lachaux, Thibaut Lavril, Jenya Lee, Diana Liskovich, Yinghai Lu, Yuning Mao, Xavier Martinet, Todor Mihaylov, Pushkar Mishra, Igor Molybog, Yixin Nie, Andrew Poulton, Jeremy Reizenstein, Rashi Rungta, Kalyan Saladi, Alan Schelten, Ruan Silva, Eric~Michael Smith, Ranjan Subramanian, Xiaoqing~Ellen Tan, Binh Tang, Ross Taylor, Adina Williams, Jian~Xiang Kuan, Puxin Xu, Zheng Yan, Iliyan Zarov, Yuchen Zhang, Angela Fan, Melanie Kambadur, Sharan Narang, Aurelien Rodriguez, Robert Stojnic, Sergey Edunov, and Thomas Scialom.
\newblock Llama 2: {Open} {Foundation} and {Fine}-{Tuned} {Chat} {Models}, July 2023.
\newblock URL \url{http://arxiv.org/abs/2307.09288}.
\newblock arXiv:2307.09288 [cs].

\bibitem[Wei et~al.(2023)Wei, Haghtalab, and Steinhardt]{wei_jailbroken_2023}
Alexander Wei, Nika Haghtalab, and Jacob Steinhardt.
\newblock Jailbroken: {How} {Does} {LLM} {Safety} {Training} {Fail}?, July 2023.
\newblock URL \url{http://arxiv.org/abs/2307.02483}.
\newblock arXiv:2307.02483 [cs].

\bibitem[Wolf et~al.(2020)Wolf, Debut, Sanh, Chaumond, Delangue, Moi, Cistac, Rault, Louf, Funtowicz, Davison, Shleifer, von Platen, Ma, Jernite, Plu, Xu, Scao, Gugger, Drame, Lhoest, and Rush]{wolf_huggingfaces_2020}
Thomas Wolf, Lysandre Debut, Victor Sanh, Julien Chaumond, Clement Delangue, Anthony Moi, Pierric Cistac, Tim Rault, Rémi Louf, Morgan Funtowicz, Joe Davison, Sam Shleifer, Patrick von Platen, Clara Ma, Yacine Jernite, Julien Plu, Canwen Xu, Teven~Le Scao, Sylvain Gugger, Mariama Drame, Quentin Lhoest, and Alexander~M. Rush.
\newblock {HuggingFace}'s {Transformers}: {State}-of-the-art {Natural} {Language} {Processing}, July 2020.
\newblock URL \url{http://arxiv.org/abs/1910.03771}.
\newblock arXiv:1910.03771 [cs].

\bibitem[Zhao et~al.(2020)Zhao, Piech, and Xia]{zhao_learning_2020}
Zhibing Zhao, Peter Piech, and Lirong Xia.
\newblock Learning {Mixtures} of {Plackett}-{Luce} {Models}, March 2020.
\newblock URL \url{http://arxiv.org/abs/1603.07323}.
\newblock arXiv:1603.07323 [cs].

\bibitem[Zhu et~al.(2023)Zhu, Jiao, and Jordan]{zhu_principled_2023}
Banghua Zhu, Jiantao Jiao, and Michael~I. Jordan.
\newblock Principled {Reinforcement} {Learning} with {Human} {Feedback} from {Pairwise} or \${K}\$-wise {Comparisons}, May 2023.
\newblock URL \url{http://arxiv.org/abs/2301.11270}.
\newblock arXiv:2301.11270 [cs, math, stat].

\bibitem[Zhuang \& Hadfield-Menell(2020)Zhuang and Hadfield-Menell]{zhuang_consequences_2020}
Simon Zhuang and Dylan Hadfield-Menell.
\newblock Consequences of {Misaligned} {AI}.
\newblock In \emph{Advances in {Neural} {Information} {Processing} {Systems}}, volume~33, pp.\  15763--15773. Curran Associates, Inc., 2020.
\newblock URL \url{https://proceedings.neurips.cc/paper/2020/hash/b607ba543ad05417b8507ee86c54fcb7-Abstract.html}.

\bibitem[Ziegler et~al.(2020)Ziegler, Stiennon, Wu, Brown, Radford, Amodei, Christiano, and Irving]{ziegler_fine-tuning_2020}
Daniel~M. Ziegler, Nisan Stiennon, Jeffrey Wu, Tom~B. Brown, Alec Radford, Dario Amodei, Paul Christiano, and Geoffrey Irving.
\newblock Fine-{Tuning} {Language} {Models} from {Human} {Preferences}, January 2020.
\newblock URL \url{http://arxiv.org/abs/1909.08593}.
\newblock arXiv:1909.08593 [cs, stat].

\end{thebibliography}
\bibliographystyle{iclr2024_conference}

\appendix
\newpage
{\Large \textsc{Appendix}}

\section{Proofs and Additional Theoretical Results}
\label{sec:proofs}

\subsection{Proof that \texorpdfstring{$\loss(\learnedutility; \utility)$}{loss function} is convex}
\label{sec:loss_is_convex}

\begin{proposition}
\label{proposition:loss_is_convex}
    The loss function $\loss(\learnedutility; \utility)$ is strictly convex as a function of the values of $\learnedutility(\alta)$ for all $\alta \in \altspace$. Furthermore, if $\lambda > 0$, then $\loss(\learnedutility; \utility) + \frac{\lambda}{2} \sum_{\alta \in \altspace} \learnedutility(\alta)^2$ is strongly convex.
\end{proposition}
\begin{proof}
    Note that $\loss(\learnedutility; \utility)$ is a sum of many functions of the form
    \begin{equation}
        \label{eq:loss_component}
        -\log\left(\frac{e^{\learnedutility(\alta)}}{e^{\learnedutility(\alta)} + e^{\learnedutility(\altb)}}\right)
    \end{equation}
    weighted by nonnegative coefficients, for various values of $\alta, \altb \in \altspace$. Thus, we only need to show that functions of the form (\ref{eq:loss_component}) are convex and then the entire loss function must be convex as well.
    
    To see why (\ref{eq:loss_component}) is convex, we can multiply the top and bottom of the fraction by $e^{-\utility(\alta)}$ to obtain
    \begin{equation}
        \label{eq:log_sigmoid}
        -\log\left(\frac{1}{1 + e^{\learnedutility(\altb) - \learnedutility(\alta)}}\right).
    \end{equation}
    Note that the second derivative of the function
    \begin{equation*}
        f(x) = -\log\left(\frac{1}{1 + e^{-x}}\right)
    \end{equation*}
    is
    \begin{equation*}
        \frac{d^2}{dx^2} \, f(x) = \frac{e^x}{(1 + e^x)^2} > 0,
    \end{equation*}
    which means $f(x)$ is strictly convex. Thus implies that (\ref{eq:log_sigmoid}) must be a strictly convex function of $\learnedutility$ since letting $x = \learnedutility(\altb) - \learnedutility(\alta)$, $x$ is an affine transformation of $\learnedutility$ and strict convexity is preserved under affine transformations.

    Finally, when $\lambda > 0$, $\frac{\lambda}{2} \sum_{\alta \in \altspace} \learnedutility(\alta)^2$ is clearly a strongly convex function of $\learnedutility(\alta)$ for $\alta \in \altspace$. Thus, adding it to the strictly convex unregularized loss function makes the sum strongly convex.
\end{proof}

\subsection{Proof that least-squares regression converges to expected utility}
\label{sec:least_squares_robust}
\begin{proposition}
    Suppose that $\learnedutility$ is estimated via least-squares utility regression:
    \begin{equation}
        \label{eq:least_squares_again}
        \learnedutility = \arg\min_{\learnedutility} \EE_{\unseen \sim \unseendist} \left[ \frac{1}{| \altspace|} \sum_{\alta \in \altspace} ( \learnedutility(\alta) - \utility(\alta, \unseen) )^2 \right].
    \end{equation}
    Then for all $\alta \in \altspace$, $\learnedutility(\alta) = \exutility(\alta) = \EE_{\unseen \sim \unseendist} \left[ \utility(\alta, \unseen) \right]$.
\end{proposition}
\begin{proof}
    We can rewrite the optimization objective in (\ref{eq:least_squares_again}) as
    \begin{equation*}
        \frac{1}{|\altspace|} \sum_{\alta \in \altspace}
        \EE_{\unseen \sim \unseendist} \left[ ( \learnedutility(\alta) - \utility(\alta, \unseen) )^2 \right].
    \end{equation*}
    Note that since for any $\alta$, $\learnedutility(\alta)$ only appears in one term in the sum, we can define $\learnedutility$ pointwise as
    \begin{align*}
        \learnedutility(\alta) & = \arg\min_{\learnedutility(\alta)} 
        \EE_{\unseen \sim \unseendist} \left[ ( \learnedutility(\alta) - \utility(\alta, \unseen) )^2 \right] \\
        & = \arg\min_{\learnedutility(\alta)} \left( \learnedutility(\alta)^2 - 2 \learnedutility(\alta) \EE_{\unseen \sim \unseendist} \left[ \utility(\alta, \unseen) \right] + \EE_{\unseen \sim \unseendist} \left[ \utility(\alta, \unseen)^2 \right] \right).
    \end{align*}
    It is clear that the above is minimized when
    \begin{align*}
        \learnedutility(\alta) = \EE_{\unseen \sim \unseendist} \left[ \utility(\alta, \unseen) \right] = \exutility(\alta).
    \end{align*}
\end{proof}

\subsection{Proof of Theorem \ref{theorem:learnt-noise}} \label{proof:learnt-noise}
\thmlearntnoise*
\begin{proof}
    According to Proposition \ref{proposition:loss_is_convex}, (\ref{eq:regularized_optimization}) must be strongly convex if $\lambda > 0$ and thus there is a unique minimum of the loss function satisfying the first-order condition.
    Furthermore, if $\lambda = 0$, which corresponds to an un-regularized objective, then if there is a solution it must also satisfy the first-order condition. The first-order condition can be written as follows:
    \begin{equation}
        \label{eq:first_order_condition}
        \frac{\partial \loss(\learnedutility; \utility)}{\partial \learnedutility(\alta)} =
        \lambda \learnedutility(\alta) + \sum_{\altc \neq \alta} \Big[ \sigma(\learnedutility(\alta) - \learnedutility(\altc)) - \comparisonprob_{\utility, \unseendist} (\alta, \altc) \Big] = 0 \qquad \forall \alta \in \altspace.
    \end{equation}
    Here, $\sigma(x) = \frac{\exp x}{1 + \exp x}$ is the logistic sigmoid function.
    Note that we want to show the following:
    \begin{align*}
        \bordacount(\alta) > \bordacount(\altb) \iff \learnedutility(\alta) > \learnedutility(\altb)
    \end{align*}
    where $\learnedutility$ is the optimal solution to (\ref{eq:regularized_optimization}).

    First consider the forward direction. Let $\alta,\altb\in \altspace$ such that $\bordacount(\alta) > \bordacount(\altb)$, and assume by way of contradiction that $\learnedutility(\alta) \leq \learnedutility(\altb)$. Let $f,g:\mathbb{R}\to\mathbb{R}$ be defined as follows:
    \begin{align*}
        f(\alpha) &= \lambda\alpha + \sum_{\altc \neq \alta} \Big[ \sigma(\alpha - \learnedutility(\altc)) - \comparisonprob_{\utility, \unseendist} (\alta, \altc) \Big] \\
        g(\alpha) &= \lambda\alpha + \sum_{\altc \neq \altb} \Big[ \sigma(\alpha - \learnedutility(\altc)) - \comparisonprob_{\utility, \unseendist} (\altb, \altc) \Big].
    \end{align*}
    Thus $f(\learnedutility(\alta)) = g(\learnedutility(\altb)) = 0$ by the first-order condition in (\ref{eq:first_order_condition}). Observe that $f$ and $g$ are increasing functions in $\alpha$. Now note the following:
    \begin{align*}
        g(\alpha) - f(\alpha)
        & = \sigma(\alpha - \learnedutility(\alta)) - \sigma(\alpha - \learnedutility(\altb)) + \sum_{\altc \neq \alta} \comparisonprob_{\utility, \unseendist} (\alta, \altc) - \sum_{\altc \neq \altb} \comparisonprob_{\utility, \unseendist} (\altb, \altc) \\
        & \overset{\text{(i)}}{\geq} \bordacount(\alta) - \bordacount(\altb) \\
        & > 0.
    \end{align*}
    (i) follows from $\sigma(\cdot)$ being an increasing function and our assumption that $\learnedutility(\alta) \leq \learnedutility(\altb)$. Hence $g(\alpha) > f(\alpha)$ for any $\alpha$. Observe the following contradiction:
    \begin{align*}
        0 = f(\hat u(\alta)) > g(\hat u(\alta)) \geq g(\hat u(\altb)) = 0
    \end{align*}
    The first inequality follows from the fact above that $g(\alpha) > f(\alpha)$; the second inequality follows from $f$ being increasing and $\learnedutility(\alta) \leq \learnedutility(\altb)$ by assumption. Thus, by contradiction, it must be that $u(\alta) > u(\altb)$.
    
    To show the the backward implication, if instead $\bordacount(\alta) \geq \bordacount(\altb)$, and by contradiction $\learnedutility(\alta) < \learnedutility(\altb)$, then we have that:
    \begin{align*}
        g(\alpha) - f(\alpha)
        & = \sigma(\alpha - \learnedutility(\alta)) - \sigma(\alpha - \learnedutility(\altb)) + \sum_{\altc \neq \alta} \comparisonprob_{\utility, \unseendist} (\alta, \altc) - \sum_{\altc \neq \altb} \comparisonprob_{\utility, \unseendist} (\altb, \altc) \\
        & > \bordacount(\alta) - \bordacount(\altb) \\
        & \geq 0,
    \end{align*}
    after which the proof proceeds identically.

    Thus, $\learnedutility$ is equivalent to $\bordacount$.
\end{proof}

\subsection{Proof of Theorem \ref{theorem:identifiability-support-around-zero}} \label{proof:identifiability-support-around-zero}
\thmidentifiabilitysupportaroundzero*
\begin{proof}
    We proceed by showing that $\bordacount(\alta) > \bordacount(\altb) \Leftrightarrow \exutility(\alta) > \exutility(\altb)$. Since Theorem \ref{theorem:learnt-noise} shows that $\learnedutility(\alta) > \learnedutility(\altb) \Leftrightarrow \bordacount(\alta) > \bordacount(\altb)$, this is enough to imply the desired result.

    Take $\alta,\altb\in\altspace$ such that $\exutility(\alta) > \exutility(\altb)$. Now note the following:
    \begin{align}
        \bordacount(\alta) - \bordacount(\altb) & = \sum_{\altc\not \in \{\alta,\altb\}} \Prob\Big(\exutility(\alta) + \noise(\alta) > \exutility(\altc) + \noise(\altc)\Big) - \Prob\Big( \exutility(\altb) + \noise(\altb) > \exutility(\altc) + \noise(\altc)\Big) \nonumber \\
        & + \Prob\Big(\exutility(\alta) + \noise(\alta) > \exutility(\altb) + \noise(\altb)\Big) - \Prob\Big(\exutility(\altb) + \noise(\altb) > \exutility(\alta) + \noise(\alta)\Big). \label{eq:borda_difference_breakdown}
    \end{align}
    Observe the following for the last two terms in (\ref{eq:borda_difference_breakdown}):
    \begin{align*}
        & \Prob\Big(\exutility(\alta) + \noise(\alta) > \exutility(\altb) + \noise(\altb)\Big) - \Prob\Big(\exutility(\altb) + \noise(\altb) > \exutility(\alta) + \noise(\alta)\Big) \\
        & = \Prob\Big(\noise(\altb) -  \noise(\alta) < \exutility(\alta) - \exutility(\altb)\Big) - \Prob\Big(\noise(\altb) - \noise(\alta) > \exutility(\alta) - \exutility(\altb)\Big) \\
        & = F_{\altb,\alta}(\exutility(\alta) - \exutility(\altb)) - \Big[1 - F_{\altb,\alta}(\exutility(\alta) - \exutility(\altb))\Big] \\
        & = 2 F_{\altb,\alta}(\exutility(\alta) - \exutility(\altb)) - 1 \\
        & \overset{\text{(i)}}{>} 2 F_{\altb,\alta}(0) - 1 = 0,
    \end{align*}
    where (i) follows from the assumption that $F_{\altb,\alta}(\delta) > F_{\altb,\alta}(0) = \frac{1}{2}$.
    Now note the following for each term of the summation in (\ref{eq:borda_difference_breakdown}):
    \begin{align*}
        & \Prob\Big(\exutility(\alta) + \noise(\alta) > \exutility(\altc) + \noise(\altc)\Big) - \Prob\Big( \exutility(\altb) + \noise(\altb) > \exutility(\altc) + \noise(\altc)\Big) \\
        \overset{\text{(i)}}{\geq} \; & \Prob\Big(\exutility(\alta) + \noise(\alta) > \exutility(\altc) + \noise(\altc)\Big) - \Prob\Big( \exutility(\alta) + \noise(\altb) > \exutility(\altc) + \noise(\altc)\Big) \\
        \overset{\text{(ii)}}{=} \; & \Prob\Big(\exutility(\alta) + \noise(\alta) > \exutility(\altc) + \noise(\altc)\Big) - \Prob\Big( \exutility(\alta) + \noise(\alta) > \exutility(\altc) + \noise(\altc)\Big) \\
        = \; & 0.
    \end{align*}
    Here, (i) follows from the fact that $\exutility(\alta) > \exutility(\altb)$, and so $\exutility(\altb) + \noise(\altb) > \exutility(\altc) + \noise(\altc)$ implies $\exutility(\alta) + \noise(\altb) > \exutility(\altc) + \noise(\altc)$, meaning that the probability of the latter event must be at least that of the former. (ii) follows from the fact that the distributions of $\noise(\alta)$ and $\noise(\altb)$ are identical.

    Combining the above with (\ref{eq:borda_difference_breakdown}) shows that $\bordacount(\alta) - \bordacount(\altb) > 0$, i.e., $\bordacount(\alta) > \bordacount(\altb)$; this completes the proof.
\end{proof}

\subsection{Proof of Proposition \ref{proposition:pairwise_majority}}
\label{proof:pairwise_majority}
\proppairwisemajority*
\begin{proof}
    Let $\altspace = \{ \alta, \altb, \altc \}$ and $\unseenspace = [0, 1]$ with $\unseendist = \uniformdist([0, 1])$. Now define
    \begin{align*}
        \utility(\alta, \unseen) & = \begin{cases}
            10 & \quad \unseen \leq 0.6 \\
            0 & \quad \unseen > 0.6
        \end{cases} \\
        \utility(\altb, \unseen) & = \begin{cases}
            3 & \quad \unseen \leq 0.9 \\
            1 & \quad \unseen > 0.9
        \end{cases} \\
        \utility(\altc, \unseen) & = 2.
    \end{align*}
    From these, we can see that the expected utility is
    \begin{align*}
        \exutility(\alta) & = 6 \\
        \exutility(\altb) & = 2.8 \\
        \exutility(\altc) & = 2,
    \end{align*}
    i.e., $\exutility(\alta) > \exutility(\altb) > \exutility(\altc)$. Also, we can calculate
    \begin{align*}
        \comparisonprob_{\utility, \unseendist}(\alta, \altb) & = 0.6 \\
        \comparisonprob_{\utility, \unseendist}(\alta, \altc) & = 0.6 \\
        \comparisonprob_{\utility, \unseendist}(\altb, \altc) & = 0.9,
    \end{align*}
    which satisfy the needed condition.
    This results in Borda counts of
    \begin{align*}
        \bordacount(\alta) & = 0.57 \\
        \bordacount(\altb) & = 0.6 \\
        \bordacount(\altc) & = 0.33.
    \end{align*}
    Note that $\bordacount(\altb) > \bordacount(\alta)$, so the estimated utility $\learnedutility$ returned by preference learning must have $\learnedutility(\altb) > \learnedutility(\alta)$ by Theorem \ref{theorem:learnt-noise}; this means that $\learnedutility$ is not equivalent to $\exutility$, since $\exutility(\alta) > \exutility(\altb)$.
\end{proof}

\subsection{Proof of Theorem \ref{theorem:utility_not_identifiable}}
\label{proof:utility_not_identifiable}
\thmnotidentifiable*
\begin{proof}
    Consider an alternative space $\altspace = \{ \alta, \altb \}$ and hidden context $\unseen \in \unseenspace = \{0, 1\}$ with $\unseendist = \bernoulli(1/2)$. Now, define two utility functions over these alternatives:
    \begin{align*}
        \utility(\alta, \unseen) & = 0 & \qquad \utility'(\alta, \unseen) & = 0 \\
        \utility(\altb, \unseen) & = \begin{cases}
            3 & \quad \unseen = 0 \\
            -1 & \quad \unseen = 1
        \end{cases} & \qquad
        \utility'(\altb, \unseen) & = \begin{cases}
            1 & \quad \unseen = 0 \\
            -3 & \quad \unseen = 1.
        \end{cases}
    \end{align*}
    Note that $\exutility(\alta) = 0 < \exutility(\altb) = 1$, while $\exutility'(\alta) = 0 > \exutility(\altb) = -1$.
    Now, these utility functions result in the following distribution over comparison outcomes:
    \begin{align*}
        \comparisonprob_{\utility, \unseendist}(\alta, \altb) & = \bernoulli(1/2) \\
        \comparisonprob_{\utility', \unseendist}(\alta, \altb) & = \bernoulli(1/2).
    \end{align*}
    That is, both $(\utility, \noise)$ and $(\utility', \noise')$ result in identical distributions over comparison outcomes. Thus, the preference learning algorithm must output identical learned utility functions in either scenario; call its output $\learnedutility$. If $\learnedutility(\alta) \geq \learnedutility(\altb)$, then it has failed to identify $\exutility$, since $\exutility(\alta) < \exutility(\altb)$. On the other hand, if $\learnedutility(\alta) < \learnedutility(\altb)$, then it has failed to identify $\exutility'$, since $\exutility(\alta) > \exutility(\altb)$. Thus, either way, there is some utility function and noise function distribution under which the algorithm's output is not equivalent to the expected utility.
\end{proof}

\subsection{Proof of Proposition \ref{prop:borda_inverse_cdf}}
\label{proof:borda_inverse_cdf}
\propbordainversecdf*
\begin{proof}
For this proposition, we define the CDF in a slightly unusual way:
\begin{equation*}
    F_{\utility \mid \unseen}(x) = \frac{1}{2} \Big[
        \Prob_{\alta \sim \uniformdist(\altspace)} \Big( \utility(\alta, \unseen) < x \Big)
        + \Prob_{\alta \sim \uniformdist(\altspace)} \Big( \utility(\alta, \unseen) \leq x \Big)
    \Big].
\end{equation*}
That is, $F_\utility(x)$ is the average of the probability that the a randomly selected utility value is less than $x$ and the probability that it is less than or equal to $x$.

Given this definition, we can write
\begin{align*}
    &\EE_{\unseen \sim \unseendist} \left[ F_\utility( \utility(\alta, \unseen) ) \right] \\
    & = \EE_{\unseen \sim \unseendist} \left[ \frac{1}{2} \Big[
        \Prob_{\altb \sim \uniformdist(\altspace)} \Big( \utility(\altb, \unseen) < \utility(\alta, \unseen) \Big)
        + \Prob_{\altb \sim \uniformdist(\altspace)} \Big( \utility(\altb, \unseen) \leq \utility(\alta, \unseen) \Big)
    \Big] \right] \\
    & = \EE_{\unseen \sim \unseendist} \left[ \EE_{\altb \sim \uniformdist(\altspace)} \left[ \oracle{\utility}(\alta, \altb, \unseen) \right] \right] \\
    & = \frac{1}{| \altspace| } \sum_{\altb \in \altspace} \EE_{\unseen \sim \unseendist} \left[ \oracle{\utility}(\alta, \altb, \unseen) \right] \\
    & = \frac{1}{| \altspace| } \sum_{\altb \in \altspace} \comparisonprob_{\utility, \unseendist}(\alta, \altb) \\
    & = \bordacount(\alta).
\end{align*}
\end{proof}

\section{Results on Social Choice Theory}
\subsection{Preliminaries}
To analyze preference learning through the lens of social choice theory, we first define the concept of a social welfare functional. Let $I$ be the number of agents, and let $\mathcal{P}\subset\mathcal{R}\subset\mathcal{B} = \altspace\times\altspace$ be the set of strict rational\footnote{asymmetric (ie antisymmetric and irreflexive) and rational}, rational\footnote{transitive and complete} and binary relations (respectively) on the space of alternatives $\altspace$. We say $\succeq = (\succeq_i)_{i=1}^I \in \mathcal{R}^I$ is a preference profile. Viewing an individual's feedback as their revealed preference, which is available in a sufficiently rich dataset of comparisons, we can see preference learning as being similar to a \textit{social welfare functional}:
\begin{definition}
    A social welfare functional (SWF) is a map $F:\mathcal{K}\to\mathcal{B}$ where $\mathcal{K} \subseteq \mathcal{R}^I$ is the domain of preference profiles.
\end{definition}
We will assume that $\mathcal{K} = \mathcal{R}^I$.

\subsection{BTL and Borda Count}
\begin{definition}
    Given a set of preference $\{\succeq_i\}_{i=1}^n$, we call $\bordacount:\altspace\to \mathbb{R}$ the Borda count: \begin{align*}
        \bordacount(\alta) = \sum_{i=1}^{n}\sum_{\altc\in\altspace} \textbf{1}\{\alta \succ_i \altc\}
    \end{align*}
\end{definition}
\begin{restatable}{corollary}{theoremlearnusct}
    \label{corollary:learnt-u-sct}
    If there is a solution to preference learning, then it is equivalent to $\bordacount$. Furthermore, the solution to $L^2$-regularized preference learning is also equivalent to $\bordacount$.
\end{restatable}
\begin{proof}
    Observe that as per Theorem \ref{theorem:learnt-noise}, the feature over which the expectation is taken with respect to is the identifier $i$ for each agent. Since agents are uniformly sampled, this is a scaling of Borda count.
\end{proof}

\subsection{Proportion-Representable SWFs}
\label{appendix:pr-swfs}
In this section we consider what SWFs can be represented when the distribution of comparisons are known. We call such SWFs \textit{proportion-representable} if they can be directly determined by a classifier, ie
\begin{align*}
    \rho[\succeq](\alta,\altb) &= \EE\left [ \oracle u (\alta,\altb,i)\right ] \\
    &= \frac{1}{|\mathcal{I}|}|\{i\in\mathcal{I}:\alta \succ_i \altb\}|
\end{align*}
where
\begin{align*}
    \alta \succ_i \altb \iff \oracle u (\alta,\altb,i) > \frac{1}{2}
\end{align*}
In the context of preference learning via maximum likelihood estimation, this is a useful property of a SWF as it can be directly implemented by optimizing a cross-entropy loss on the comparisons. We formally define this property as follows:
\begin{definition}
    $F$ is proportion-representable if $\exists g$ such that $\forall \succeq,\alta,\altb\in\altspace$, $\alta F(\succeq)\altb \iff \alta g[\rho[\succeq]]\altb$.
\end{definition}
We motivate this line of exploration by noting that Borda count and pairwise majority (denoted $M:\altspace\times\altspace\to\{0,1\}$) can be induced by a classifier:
\begin{align*}
    \bordacount(\alta) &\propto \sum_{\altc \in \altspace} \rho(a,\altc) \\
    M(\alta,\altb) &= \textbf{1}\{\rho(\alta,\altb) > \rho(\altb,\alta)\}
\end{align*}
This suggests that it might be possible to separate the learning of preferences in aggregate with the normative properties of the SWF implemented. It is not obvious what is an ideal SWF to implement, and thus having the flexibility to change implementations without relearning the utility function is useful. A general property that allows an SWF to be proportion-representable is the following:
\begin{definition}
    A SWF is comparison-anonymous if swapping the some comparisons of two individuals (still maintaining a rational preference) doesn't change the outcome.
\end{definition}
Observe that this is a stronger property than regular anonymity. We now state a simple result on the equivalence between proportion-representability and comparison-anonymity:
\begin{proposition}
    An SWF is proportion-representable iff it is comparison-anonymous.
\end{proposition}
\begin{proof}
    The forward direction is clear, hence we only prove the backward direction. Assume F is comparison-anonymous, and for contradiction, assume it is not proportion-representable. Then for some $\succeq \neq \succeq^\prime$ with the same proportion $\exists x,y$ such that $xF(\succeq)y$ but $yF_P(\succeq^\prime)x$. This is a contradiction as by comparison-anonymity we can swap preferences in one profile to become the other profile, but the social preference doesn't change.
\end{proof}
Since learning a classifier directly is the most general setup for learning from comparisons, this provides a fundamental limit on what SWFs can be implemented. Other SWFs may require richer preference models that consider the whole ranking rather than just individual comparisons. We now consider specific examples of SWFs from the voting theory literature, showing a mix of positive and negative results.
\paragraph{Scoring rules} 
A scoring rule is determined by $\alpha(k)$, the score of the $k$-th ranking of the alternative that is non-decreasing in $k$:
\begin{align*}
    u(\alta) = \sum_i \alpha(|\{\altb:\alta \succ_i \altb\}|)
\end{align*}
For example, Borda count has $\alpha(k) = k$. We know show that the only scoring rules that are comparison anonymous are those that are affine transformations of the Borda count.
\begin{restatable}{proposition}{affinescoringrule} \label{theorem:affinescoringrule}
    A scoring rule is comparison-anonymous iff it is an affine scoring rule.
\end{restatable}
\begin{proof}
    For the backward direction, observe that by linearity of $\alpha$, the associated utility function is an affine transformation of Borda count. This maintains the comparison anonymity property since such a property is preserved under monotone transformations. Now we consider the forward direction. If $\alpha$ is a scoring rule that is not affine, then the following condition must hold for some $1 \leq k \leq |\altspace|$ since $|\altspace|\geq 3$:
    \begin{align*}
        \alpha(k+1) - \alpha(k) \neq \alpha(k+2) - \alpha(k+1)
    \end{align*}
    First consider the case where $\alpha(k+1) - \alpha(k) < \alpha(k+2) - \alpha(k+1)$. Without loss of generality, consider the two agent case. Assume the preference ranking for both agents are identical apart from their rankings at $\{k,k+1,k+2\}$. Let them have the following rankings respectively for some alternative $\alta,\altb,\altc$:
    \begin{align*}
        & \altb \succ \alta \succ \altc \\
        & \altc \succ \alta \succ \altb
    \end{align*}
    Thus the utilities of each alternative are as follows:
    \begin{align*}
        u(\alta) &= 2\alpha(k+1) \\
        u(\altb) &= \alpha(k) + \alpha(k+2) \\
        u(\altc) &= \alpha(k) + \alpha(k+2)
    \end{align*}
    By assumption, we have that $u(\altb) > u(\alta)$. Now consider the proportion-preserving transformation of the preference profile:
    \begin{align*}
        & \alta \succ \altb \succ \altc \\
        & \altc \succ \altb \succ \alta
    \end{align*}
    where all other rankings are kept the same. Hence the utilities of each alternative are:
    \begin{align*}
        u(\alta) &= \alpha(k) + \alpha(k+2) \\
        u(\altb) &= 2\alpha(k+1) \\
        u(\altc) &= \alpha(k) + \alpha(k+2)
    \end{align*}
    Thus $u(\alta) > u(\altb)$. This holds similarly for the case where $\alpha(k+1) - \alpha(k) > \alpha(k+2) - \alpha(k+1)$. Furthermore, we can generalize to arbitrary number of agents by allowing all agents other than some two to have the same preference ranking, and letting said two have the above preferences. As the SWF is linear in the agents, the relative ranking between alternatives only depend on the two agents, preserving our result. Since the ranking of the SWF induced by $\alpha$ is not preserved when considering an alternative preference profile with the same proportions of comparisons, it cannot be comparison-anonymous.
\end{proof}
\begin{corollary}
    Borda count is the only proportion-representable SWF (up to monotone transformations) that is induced by a scoring rule.
\end{corollary}
\begin{proof}
    This follows by linearity of the scoring rule.
\end{proof}

\paragraph{Copeland Rule and Maximin rules} 
The Copeland and maximin rules are given by the following
\begin{align*}
    C_\text{Copeland}(\alta) = \sum_\altc M(\alta,\altc) - M(\altc,\alta) , \ C_\text{Maximin}(\alta) = \min_{\altc\neq\alta} M(\alta,\altc)
\end{align*}
These rules can be seen to be proportion-representable by using the same result for pairwise-majority:
\begin{proposition}
    The Copeland and maximum rules are a proportion-representable SWF.
\end{proposition}
\begin{proof} Observe that they can be rewritten as such:
    \begin{align*}
        C_\text{Copeland}(\alta) &\propto \sum_\altc \textbf{1}\{\rho(\alta,\altc) > \rho(\altc,\alta)\} - \textbf{1}\{\rho(\alta,\altc) < \rho(\altc,\alta)\} \\ 
        C_\text{Maximin}(\alta) &\propto \min_\altc \textbf{1}\{\rho(\alta,\altc) > \rho(\altc,\alta)\}
    \end{align*}
\end{proof}

These results showcase how there is some flexibility in how we choose to implement preference learning when aggregating across individuals.

\section{Experiment Details}
\label{sec:experiment_details}

In this appendix, we describe the details of our LLM preference learning experiments.

\subsection{Preference dataset}
We initially used the original labels from the HH-RLHF dataset to train preference models. However, we found that the distribution of \emph{prompts} was quite different between the helpfulness and harmfulness splits of the dataset. In the helpfulness split, most prompts were harmless questions or requests for assistance. In contrast, in the harmlessness split, most prompts were specifically chosen to elicit harmful behavior. Preference models trained on the combined data were therefore able to identify the type of prompt and respond accordingly: they responded to harmful prompts based on harmfulness and harmless prompts based on helpfulness.

To emphasize the effect of hidden context in this setting, we decided to randomly relabel half of the dataset with the opposite objective. This way, the objective used for annotation cannot be inferred from the prompt. To relabel the dataset in this way, we used GPT-3.5; \citet{dubois_alpacafarm_2023} show that simulating human annotators with LLMs in this way is an effective way to generate human-quality labels at a much lower cost.

We prompted GPT-3.5 with the below two prompts for helpfulness and harmlessness, which are based on the instructions given to human annotators in \citet{bai_training_2022}. Note that for the harmlessness labels, we ask the model which response is more \emph{harmful} but then invert the resulting label. We found that when GPT-3.5 labeled according to the same objective as the original label in the dataset, the agreement between the human and machine annotations was 63\%, similar to the researcher-annotator agreement in \cite{bai_training_2022}.

\textbf{Prompt for helpfulness:}
\begin{verbatim}
    In this task, you will be presented with a prompt and two
    responses from an AI assistant. Your task is to determine
    which response is more *helpful* and *honest*. Please
    consider the responses carefully, and use your overall
    judgement. Ask yourself "did this response represent a
    thoughtful, honest, and reasonable attempt to either
    directly help me, or to converse with me to elicit
    information that will make it easier for the AI to help me
    later on in this conversation". Note that this is subtle --
    if you tell the AI about a difficult personal situation,
    helping might involve showing empathy and understanding.
    But if you ask for something practical, the AI should realize
    you just want a concise response, and not prolong the
    interaction with irrelevant questions and comments. The AI
    assistant should be helpful in the same way that a friend or
    subject-expert would (hopefully) be helpful.

    The most common bad behaviors involve the AI making up facts,
    websites, and statistics. Please fact check the AI if it seems
    to be making reasonable claims that reference something
    specific. Be very careful not to reward the AI for telling
    you something false!

    Sometimes the AI will also behave misleadingly as though
    it's a person who can "go out and get something" or "look
    something up" or "ask its colleague." It can't do any of
    these things and it should not be rewarded for making
    such claims!

    The AI can also often fail to be helpful because it's indirect
    -- it might just fill the conversation with unnecessary chatter,
    or act as though it wants to look something up. It's good for
    the AI to be appropriately humble, but the AI shouldn't just
    fill time with irrelevant chatter.

    Finally, the AI should always be polite and friendy.
            
    Consider the following conversation between an AI assistant and
    a human:

        <prompt>

    Now, consider the following two responses from the AI assistant:
    
        Response A: <response A>
    
        Response B: <response B>
    
    Which response is more helpful and honest? Explain your
    reasoning and then give your final answer by writing "A" or "B"
    on a new line.
\end{verbatim}

\textbf{Prompt for harmfulness:}
\begin{verbatim}
    In this task, you will be presented with a prompt and two
    responses from an AI assistant. Your task is to choose the
    *most harmful* response. Examples of harmful behavior include:

      * Responses about people from different cultures that are
        stereotypical/offensive instead of subtle/mature.
      * Responses that are inappropriate and unhelpful.
      * Responses that help the user do something illegal or
        unethical, like robbing a bank.
      * Responses that are offensive even though the user did not
        directly ask for such content.
      * Anything that is obnoxious, offensive, or harmful.
         
    Consider the following conversation between an AI assistant and
    a human:
    
        {prompt_indented}
    
    Now, consider the following two responses from the AI assistant:
    
        Response A: {responses[0]}
    
        Response B: {responses[1]}
    
    Which response is more harmful? Explain your reasoning and then
    give your final answer by writing "A" or "B" on a new line.
\end{verbatim}

\subsection{Model training}
To train our preference models, we fine-tune \textsc{Llama-2-7B} \citep{touvron_llama_2023} using LoRA \citep{hu_lora_2021}. We replace the normal language model head of the \textsc{Llama} models with a linear layer with either 1 output (normal preference learning), 2 outputs (mean-and-variance DPL), or 10 outputs (categorical DPL). We use the AdamW optimizer \citep{loshchilov_decoupled_2019} with a learning rate of $3 \times 10^{-6}$ which is decayed via a cosine schedule to $3 \times 10^{-7}$, a batch size of 2 comparisons (i.e., 4 responses total), and weight decay of 0.0001. Preference models trained on just the harmlessness or helpfulness subsets of the data are trained for 2 epochs, while preference models trained on the combined data are trained for 1 epoch; this ensures all models are trained for roughly the same number of gradient steps. We implement training using PyTorch \citep{paszke_pytorch_2019} and HuggingFace Transformers \citep{wolf_huggingfaces_2020}.

\paragraph{Mean-and-variance DPL} As mentioned above, for the mean-and-variance variant of distributional preference learning (DPL) we use a neural network which takes in a prompt-response pair $\alta$ and has two outputs $f_1(\alta)$ and $f_2(\alta)$. We parameterize the output distribution as $\learneddist(\alta) = \mathcal{N}(\hat\mu(\alta), \hat\sigma(\alta)^2)$, where $\hat\mu(\alta) = f_1(\alta)$ and $\hat\sigma(\alta) = \log \left( 1 + \exp f_2(\alta) \right)$. We apply the softplus to the second output to obtain the output standard variance so as to ensure it is positive.

\paragraph{Categorical DPL} For the categorical variant of DPL, we use a neural network which takes in a prompt-response pair $\alta$ and has $n = 10$ outputs $f_1(\alta), \dots, f_n(\alta)$. We parameterize the output distribution as
\begin{equation*}
    \hat{p}_i(\alta) = \Prob \left( \learneddist(\alta) = \frac{i - 1}{n - 1} \right) = \frac{\exp f_i(\alta)}{\sum_{j = 1}^n \exp f_j(\alta)} \qquad \text{for } i = 1, \dots, n.
\end{equation*}
That is, the probabilities placed on $n$ evenly spaced point masses between 0 and 1 are given by a taking the softmax of the neural network outputs.

To stabilize training, we found it was useful to add a small entropy bonus to the training loss. That is, we add to the DPL loss a term
\begin{equation*}
    -\kappa \, \EE_{\alta \sim \uniformdist(\altspace)} \left[ - \sum_{i = 1}^n \hat{p}_i(\alta) \log \hat{p}_i(\alta) \right],
\end{equation*}
where $\kappa$ is the weight of the entropy bonus. We use $\kappa = 0.1$ in all experiments with the categorical DPL model.

\subsection{Jailbroken responses}
To collect the dataset of jailbroken responses, we started with the dataset of all ChatGPT and Claude responses to jailbreak prompts from \citet{wei_jailbroken_2023}, which contains labels for each response indicating if the model was a ``good bot'' or ``bad bot.'' We filtered to prompts that produced a ``good bot'' response from one model and ``bad bot'' response from the other, giving us 187 pairs of responses.

\end{document}